\documentclass[10pt,twocolumn,letterpaper]{article}

\usepackage{algorithm,algpseudocode}
\usepackage{preprint}
\usepackage[T1]{fontenc}

\usepackage{amsmath,amsfonts,bm}

\def\eqref#1{equation~\ref{#1}}

\def\1{\bm{1}}

\DeclareMathAlphabet{\mathsfit}{\encodingdefault}{\sfdefault}{m}{sl}
\SetMathAlphabet{\mathsfit}{bold}{\encodingdefault}{\sfdefault}{bx}{n}

\usepackage{url,array,colortbl,tabularx,multirow}
\usepackage{wrapfig,placeins}
\definecolor{fraInk}{HTML}{034ED6}
\definecolor{fraTint}{HTML}{FEFAF3}
\definecolor{deltaRed}{HTML}{AA4C45}
\definecolor{deltaGreen}{HTML}{3F8F4E}
\definecolor{oursLav}{HTML}{E4EDF8}
\definecolor{rowTint}{HTML}{E4EDF8}
\definecolor{linkblue}{rgb}{0.21,0.49,0.74}
\usepackage[pagebackref,breaklinks,colorlinks,allcolors=linkblue]{hyperref}
\usepackage{iftex}
\ifXeTeX\fi

\title{Forensic Reserve: Localizing and Eliciting Latent Knowledge \\ for Image Forgery Detection}
\author{Jiahua Li$^{1}$, Zixu John$^{2}$, Tom Zhong$^{2}$, Fuping Wu$^{3}$,\\
Tianhao Xu$^{2}$, Jianqing Zheng$^{1}$, Yuanhan Mo$^{3}$, Fei Shen$^{4,*}$\\
{\small $^{1}$University of Oxford, the United Kingdom\quad $^{2}$Independent Researcher}\\
{\small $^{3}$Imperial College London, the United Kingdom}\\
{\small $^{4}$National University of Singapore, Singapore\quad $^{*}$Corresponding author.}}
\hypersetup{pdftitle={Forensic Reserve: Localizing and Eliciting Latent Knowledge for Image Forgery Detection},pdfauthor={Jiahua Li, Zixu John, Tom Zhong, Fuping Wu, Tianhao Xu, Jianqing Zheng, Yuanhan Mo, Fei Shen}}

\makeatletter
\newif\ifshiftfirstcolumn\shiftfirstcolumntrue
\AddToHook{cmd/@outputdblcol/before}{%
  \ifshiftfirstcolumn
    \if@firstcolumn\else
      \global\shiftfirstcolumnfalse
      \setbox\@outputbox\vbox to\ht\@outputbox{%
        \kern6.4238pt\box\@outputbox\vss}%
    \fi
  \fi}
\makeatother
\begin{document}
\raggedbottom
\maketitle
\begin{abstract}

As generated images become increasingly realistic, reliable forgery detection is essential for maintaining trust in visual information.
However, existing methods primarily rely on task-specific supervision to adapt Vision Foundation Model representations, without fully exploiting internal forensic knowledge to guide detection.
To address this limitation, we propose Reserve-Guided Elicitation (RGE), a framework that treats sparse, origin-sensitive internal components in pretrained models as a forensic reserve and translates their localization into structural constraints for lightweight adaptation.
Specifically, we first use the Forensic Lens (F-lens) to decompose activations across layers and token groups into independent components and globally screen them by their response differences between real and fake images, identifying reserve sites and directions.
Next, we map the selected directions back to hidden-state space to construct fixed reserve subspaces and insert a Forensic Reserve Adapter (FRA) at each identified site.
Finally, with the backbone parameters, previously fitted reference classifier, and subspace bases fixed, we train only the FRA coefficient maps to generate input-dependent residual updates constrained to the corresponding subspaces, strengthening existing forensic responses.
Using 5,000 labeled training images and a trainable parameter budget below 0.2\% of the backbone, RGE achieves competitive performance across three detection benchmarks without target-benchmark adaptation.
Furthermore, RGE consistently improves over the corresponding frozen detectors across eight encoders spanning self-supervised and vision-language pretraining, eliciting a latent forensic capacity broadly shared across pretrained vision models.

\end{abstract}

\section{Introduction}
Recent studies~\citep{yan2025aide,li2025aigibench} show that as generated images become increasingly realistic, distinguishing them from real images becomes more challenging.
Maintaining trust in visual media therefore calls for dependable forgery detection.
Practical detectors must also generalize to evolving image generators while keeping the costs of both data annotation and model training at manageable levels.

\begin{figure}[t]
    \centering
    \includegraphics[width=0.98\linewidth]{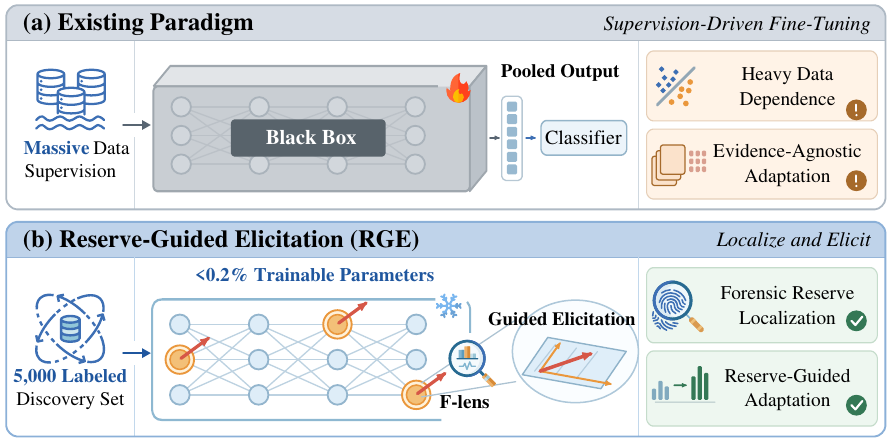}
\vspace{-0.3cm}
    \caption{\textbf{Comparison of Forgery Detection Paradigms.}
    Fine-tuning relies on large labeled sets without localizing forensic evidence; RGE localizes the forensic reserve to guide targeted enhancement of existing internal responses.}
    \label{fig:rge-concept}
\vspace{-12pt} %
\end{figure}

Existing methods address image forgery detection through explicit artifact analysis~\citep{li2025safe,wang2023dire} and adaptation of pretrained representations~\citep{guillaro2025bfree,chen2025dda,yan2026dgsnet}.
Artifact-based detectors exploit spatial and frequency statistics~\citep{li2025safe} or diffusion reconstruction residuals~\citep{wang2023dire}, but their effectiveness depends on whether these cues remain discriminative across generation settings.
Methods based on Vision Foundation Models (VFMs) adapt pretrained representations through forgery supervision, refining training data~\citep{guillaro2025bfree,chen2025dda} and parameter updates~\citep{hu2022lora,yan2026dgsnet} to improve generalization.
As illustrated in Figure~\ref{fig:rge-concept}~(a), these approaches primarily optimize real--fake separation through classification objectives.
However, improved classification does not directly reveal which internal components carry forensic information or specify where and along which directions adaptation should act.
The locations and directions of internal forensic knowledge therefore remain underutilized as structural guidance for targeted enhancement of VFM-based forgery detectors.

Recent studies~\citep{koutlis2024rine,dou2026dna} show that forensic information is unevenly distributed across internal features.
This motivates our hypothesis that components carrying image-origin evidence can provide discriminative features and structural priors for adaptation.
Drawing inspiration from cognitive reserve~\citep{stern2002reserve}, we refer to sparse, origin-sensitive internal components as a forensic reserve.
Once localized, their sites indicate where adaptation should act, while their mapped directions define the subspaces available for residual updates (Figure~\ref{fig:rge-concept}~(b)).
This structure offers a way to focus limited supervision on enhancing existing forensic responses.

For that, we propose Reserve-Guided Elicitation (RGE), a framework that translates reserve localization into structural constraints for lightweight adaptation.
Specifically, the Forensic Lens (F-lens) decomposes activations across layers and token groups into independent components and globally screens them by their response differences between real and fake images, identifying reserve sites and directions.
We map the selected directions back into the corresponding hidden-state spaces to construct fixed reserve subspaces and insert a Forensic Reserve Adapter (FRA) at each identified site.
Each FRA reads the current hidden state to produce an input-dependent residual update constrained to its reserve subspace.
The backbone parameters, previously fitted reference classifier, and subspace bases remain fixed; only the FRA coefficient maps are trained, concentrating limited forgery supervision on the targeted enhancement of existing forensic responses. Our main contributions are summarized as follows:
\begin{itemize}
    \item We propose RGE, a lightweight adaptation framework for forgery detection. It localizes a forensic reserve in pretrained vision models and recruits its sparse, origin-sensitive internal components for targeted enhancement.

    \item We develop F-lens and FRA to connect reserve localization with constrained residual updates. Site ablations and residual-update interventions support the functional contribution of the identified sites and learned residual updates to detection performance.

    \item RGE achieves competitive detection across three main benchmarks without target-benchmark adaptation, using 5,000 labeled training images and trainable parameters below 0.2\% of the backbone. Further evaluations show consistent gains over frozen baselines across eight Vision Transformer (ViT) encoders from four model families.
\end{itemize}

\section{Related Work}

\noindent\textbf{AI-Generated Image Detection.}
Existing detectors identify AI-generated images through artifact analysis, frozen feature readout, or adaptation of VFMs.
Artifact-based methods exploit spatial and frequency signatures, as in SAFE~\citep{li2025safe}, or diffusion reconstruction residuals, as in DIRE~\citep{wang2023dire}, although these cues become less discriminative as generated images grow more realistic~\citep{yan2025aide,li2025aigibench}.
Frozen readout methods use CLIP features~\citep{ojha2023univfd,cozzolino2024raising} or perturbation responses~\citep{he2024rigid}, while RINE~\citep{koutlis2024rine} accesses intermediate-block representations.
For VFM adaptation, DRCT~\citep{chen2024drct}, B-Free~\citep{guillaro2025bfree}, and DDA~\citep{chen2025dda} refine supervision through reconstruction-based hard negatives, content matching, and pixel- and frequency-aligned image pairs, respectively.
DGS-Net~\citep{yan2026dgsnet} and Effort~\citep{yan2025effort} constrain parameter updates through gradient surgery and pretrained weight subspaces, respectively.
At the level of internal features, DNA~\citep{dou2026dna} localizes sparse forgery-discriminative units and validates their relevance through masking.
However, these methods offer limited guidance on using localized forensic features to determine adaptation sites and update subspaces.

\noindent\textbf{Latent Features and Targeted Adaptation.}
Research on latent features spans representation decomposition, activation steering, and feature-guided fine-tuning.
For representation decomposition, sparse autoencoders (SAEs)~\citep{cunningham2024sae} learn sparse dictionaries from activations, while ICA Lens~\citep{liu2026icalens,hyvarinen1999fastica,hyvarinen2000ica} provides independent component coordinates with associated reading and writing maps.
For activation steering, J-lens~\citep{gurnee2026verbalizable} and TraceRouter~\citep{shi2026tracerouter} control model behavior through concept-coordinate swaps and activation rescaling along identified propagation paths, respectively.
In vision, feature suppression~\citep{joseph2025steeringclip} improves CLIP robustness, while sparse steering~\citep{chatzoudis2026visualsteering} examines the benefits of feature-level edits and the limitations of reconstruction-based feature selection.
Beyond prescribed activation edits, SAE-FT~\citep{morelli2026saeft} uses sparse feature structure to regularize fine-tuning of both the visual encoder and the classifier.
However, learning input-dependent updates within fixed feature subspaces while keeping both the backbone and the classifier frozen remains underexplored.

\begin{figure*}[t]
\centering
\includegraphics[width=0.98\linewidth]{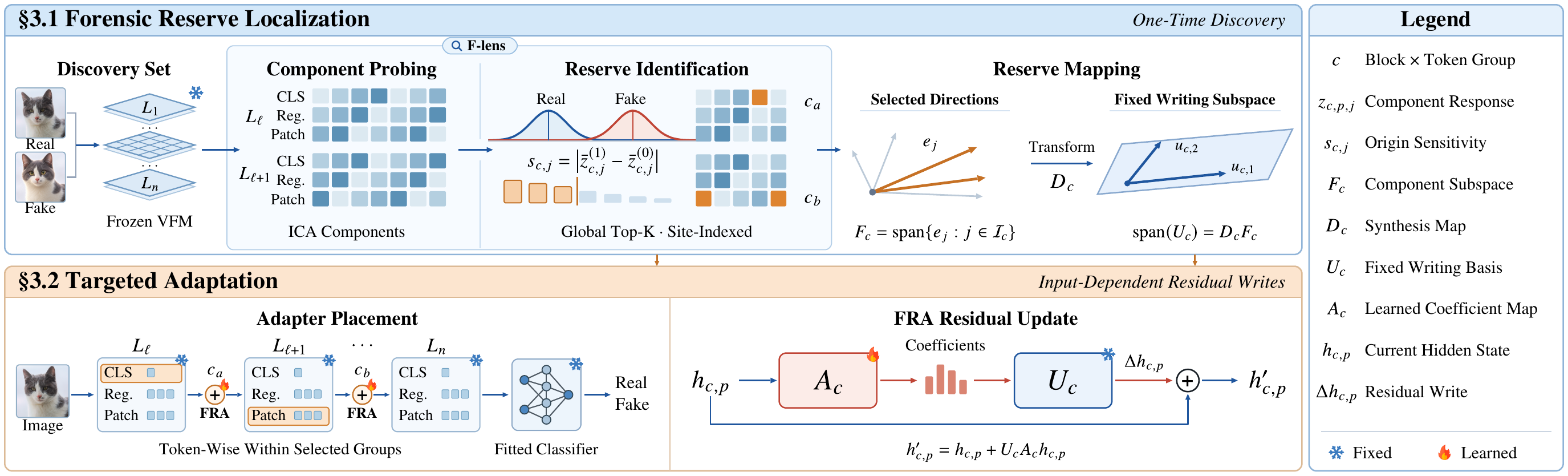}
\vspace{-0.3cm}
\caption{\textbf{Overview of Reserve-Guided Elicitation (RGE).} F-lens localizes the forensic reserve in the frozen backbone (Section~\ref{sec:reserve-localization}); FRA writes within the mapped reserve subspaces at selected sites (Section~\ref{sec:reserve-elicitation}).}
\label{fig:loop}
\vspace{-9pt} %
\end{figure*}

\section{Method}
\label{sec:method}
RGE uses origin-sensitive components in a pretrained vision backbone as structural priors for targeted adaptation.
Given a labeled discovery set, we first fit a linear classifier to the backbone's frozen pooled representations and select its decision threshold on a held-out development split, forming a reference detector.
RGE then proceeds in two stages (Figure~\ref{fig:loop} and Algorithm~\ref{alg:rge}).
In forensic reserve localization (upper row), F-lens decomposes internal activations from the unadapted backbone and selects components with strong real--fake response contrasts.
Their locations determine the adaptation sites, while their mapped directions define fixed writing subspaces in hidden-state coordinates.
In targeted adaptation (lower row), FRA inserts residual branches at these sites and learns coefficient maps that generate input-dependent residual writes within the corresponding subspaces.
Only these coefficient maps are trained with detection supervision; the backbone, reference classifier, and writing bases remain fixed.

\begin{algorithm}[t]
\caption{Reserve-Guided Elicitation (RGE)}
\label{alg:rge}
\small
\begin{algorithmic}[1]
\Require Frozen backbone $\theta$; fitting set $\mathcal D$;
development set $\mathcal D_{\mathrm{dev}}$; budget $K$;
training steps $T$.
\Ensure Adapted detector $f_{\mathcal A}$.
\vspace{2.691075pt}

\State Fit a weighted classifier on standardized frozen features
from $\mathcal D$; select $\tau$ on $\mathcal D_{\mathrm{dev}}$
to form $\psi$.

\vspace{2.691075pt}
\Statex \textbf{Forensic Reserve Localization}
\ForAll{sites $c=(\ell,t)$}
    \State Fit ICA to normalized activations from $\mathcal D$;
    obtain $R_c$ and unit-variance coordinates $z_{c,p}$.
    \State Compute $s_{c,j}$ using Eq.~(\ref{eq:origin-sensitivity}).
\EndFor
\State Select global top-$K$ components to form
$\mathcal I_c$ and $\mathcal C_\star$.
\State Construct $U_c$ at each selected site
using Eq.~(\ref{eq:reserve-writing-space}).

\vspace{2.691075pt}
\Statex \textbf{Targeted Adaptation}
\State Insert FRA at $\mathcal C_\star$; initialize all $A_c=0$.
\State Fix $\theta$, $\psi$, $\{U_c\}$, and feature standardization.
\For{$t=1,\ldots,T$}
    \State Apply FRA updates Eq.~(\ref{eq:fra-update})
    and compute scores Eq.~(\ref{eq:detail-detector-score}).
    \State Update only $\mathcal A$
    using the weighted BCE loss Eq.~(\ref{eq:detail-detection-loss}).
\EndFor
\vspace{2.691075pt}
\State \Return $f_{\mathcal A}$;
predict fake when $f_{\mathcal A}(x)\geq 0$.
\end{algorithmic}
\end{algorithm}

\subsection{Forensic Reserve Localization}
\label{sec:reserve-localization}

\noindent\textbf{Discovery Set.}
We use a labeled discovery set balanced between real and fake images and covering multiple real-image domains and generation or manipulation families.
Sampling prioritizes coverage across underlying sources and generator families before drawing additional examples from the same groups.
The same fitting images support reference-classifier fitting, reserve localization, and FRA training, while a separate development split is held out for threshold selection.
The discovery data are curated from public sources independently of the target benchmarks, on which the resulting detector is evaluated without further fitting.
The appendix details source curation, group-disjoint partitioning, and hierarchical sampling.

\noindent\textbf{Component Probing.}
F-lens decomposes activations separately by block and token group so that each component retains a location for subsequent adaptation.
We define a site as $c=(\ell,t)$, where $\ell$ indexes a transformer block and $t$ denotes an available token group, such as classification (CLS), register, or patch tokens.
Let $h^0_{c,p}(x)\in\mathbb R^{d_c}$ denote the unadapted hidden state of token $p$ at site $c$ for image $x$, collected at the block output.
We collect all CLS and register tokens where present and spatially sample patch tokens to form each site's activation population.
At each site, F-lens applies $\ell_2$ normalization and mean centering, followed by whitening and independent component analysis (ICA).
Each component is then scaled to unit empirical variance, yielding the final reading map $R_c$ and coordinates
\[
z_{c,p}(x)
=
R_c\left(\widehat h^0_{c,p}(x)-\mu_c\right),
\]
where $\widehat h^0_{c,p}(x)$ is the normalized hidden state and $\mu_c$ is its empirical site mean.
This standardization expresses component response contrasts in common units for global ranking.

\noindent\textbf{Reserve Identification.}
We identify reserve components through their origin sensitivity, measured by the absolute difference between their mean responses to real and fake images:
\begin{equation}
s_{c,j}
=\left|
\mathbb E\!\left[z_{c,p,j}(x)\mid y=1\right]
-\mathbb E\!\left[z_{c,p,j}(x)\mid y=0\right]
\right|,
\label{eq:origin-sensitivity}
\end{equation}
where $y=0$ and $y=1$ denote real and fake images, respectively, and the expectations are empirical averages over each class's sampled token activations.
The absolute difference captures response contrasts in either direction.
Under the orthogonal ICA rotation, $\sum_j s_{c,j}^2$ equals the squared Mahalanobis separation between the normalized class means, using the pooled empirical covariance restricted to the retained subspace.
We retain the top $K$ components globally across all fitted sites, allocating the component budget across layers and token groups according to their response contrasts.
These selected, site-indexed components constitute the forensic reserve.
Let $\mathcal I_c$ denote the selected component indices at site $c$, and let $\mathcal C_\star=\{c:\mathcal I_c\neq\varnothing\}$ specify the sites for adapter placement.
At each selected site, the corresponding coordinate axes span
$F_c=\operatorname{span}\{e_j:j\in\mathcal I_c\}$,
where $e_j$ is the $j$th coordinate basis vector.
The family $\{F_c:c\in\mathcal C_\star\}$ forms the \emph{forensic reserve space} (F-space).
The appendix provides implementation details for activation collection, ICA fitting, and component selection.

\noindent\textbf{Reserve Mapping.}
F-space is expressed in component coordinates, whereas residual updates act in hidden-state coordinates.
We connect these spaces through the synthesis map $D_c=R_c^\dagger$, where $\dagger$ denotes the Moore--Penrose pseudoinverse.
At each selected site, we orthonormalize the columns of $D_c[:,\mathcal I_c]$ to obtain a writing basis $U_c\in\mathbb R^{d_c\times r_c}$ satisfying
\begin{equation}
D_cF_c=\operatorname{span}(U_c).
\label{eq:reserve-writing-space}
\end{equation}
Here $r_c$ is the rank of the selected synthesis directions; orthonormalization changes their basis without changing their span.
The directions estimated in normalized activation space serve as fixed directional priors for residual writes in raw hidden-state space, confining each local update to $\operatorname{span}(U_c)$.
For each backbone and discovery configuration, localization is performed once.
The selected sites and writing bases remain fixed throughout adaptation and are reused across evaluation benchmarks.

\begin{table*}[t]
\centering

\caption{\textbf{Detection Across Three Benchmarks (\%).} Average is the mean over the three benchmarks. Best and second-best are bold and underlined; N/A denotes unavailable results.}
\label{tab:main-results}
\vspace{-0.3cm}
\begin{minipage}{\linewidth}
\begingroup
\fontsize{9}{10.5}\selectfont
\setlength{\tabcolsep}{1pt}
\renewcommand{\arraystretch}{1.12}
\begin{tabularx}{\linewidth}{@{}>{\hsize=1.8\hsize\linewidth=\hsize\raggedright\arraybackslash}X!{\hspace{2pt}\vrule\hspace{1pt}}>{\hsize=.9\hsize\linewidth=\hsize\centering\arraybackslash}X>{\hsize=.9\hsize\linewidth=\hsize\centering\arraybackslash}X!{\hspace{6pt}\vrule\hspace{1pt}}>{\hsize=.9\hsize\linewidth=\hsize\centering\arraybackslash}X>{\hsize=.9\hsize\linewidth=\hsize\centering\arraybackslash}X!{\hspace{6pt}\vrule\hspace{1pt}}>{\hsize=.9\hsize\linewidth=\hsize\centering\arraybackslash}X>{\hsize=.9\hsize\linewidth=\hsize\centering\arraybackslash}X!{\hspace{6pt}\vrule\hspace{1pt}}>{\hsize=.9\hsize\linewidth=\hsize\centering\arraybackslash}X>{\hsize=.9\hsize\linewidth=\hsize\centering\arraybackslash}X@{\hspace{7pt}}}
\toprule
 & \multicolumn{2}{c!{\hspace{6pt}\vrule\hspace{1pt}}}{\textbf{AIGIBench}~\citep{li2025aigibench}} & \multicolumn{2}{c!{\hspace{6pt}\vrule\hspace{1pt}}}{\textbf{Chameleon}~\citep{yan2025aide}} & \multicolumn{2}{c!{\hspace{6pt}\vrule\hspace{1pt}}}{\textbf{HiRes-50K}~\citep{mu2026hida}} & \multicolumn{2}{c@{\hspace{7pt}}}{\textbf{Average}} \\
\cmidrule(lr){2-3}\cmidrule(lr){4-5}\cmidrule(lr){6-7}\cmidrule(l){8-9}
\textbf{Method} & \multicolumn{1}{c}{\textbf{mAP} $\uparrow$} & \multicolumn{1}{c!{\hspace{6pt}\vrule\hspace{1pt}}}{\textbf{Acc} $\uparrow$} & \multicolumn{1}{c}{\textbf{mAP} $\uparrow$} & \multicolumn{1}{c!{\hspace{6pt}\vrule\hspace{1pt}}}{\textbf{Acc} $\uparrow$} & \multicolumn{1}{c}{\textbf{mAP} $\uparrow$} & \multicolumn{1}{c!{\hspace{6pt}\vrule\hspace{1pt}}}{\textbf{Acc} $\uparrow$} & \multicolumn{1}{c}{\textbf{mAP} $\uparrow$} & \multicolumn{1}{c@{\hspace{7pt}}}{\textbf{Acc} $\uparrow$} \\
\midrule
UnivFD~\citep{ojha2023univfd} & 75.60 & 72.50 & 46.20 & 57.20 & 54.93 & 62.05 & 58.91 & 63.92 \\
LTD~\citep{yang2026ltd} & 77.60 & 74.90 & 35.21 & 57.52 & 62.96 & 52.61 & 58.59 & 61.68 \\
PLM~\citep{zhou2026plm} & 81.10 & 78.80 & 46.47 & 59.36 & 49.90 & 49.94 & 59.16 & 62.70 \\
AIDE~\citep{yan2025aide} & 82.70 & 77.60 & 69.70 & 65.80 & 74.61 & 56.46 & 75.67 & 66.62 \\
DRCT~\citep{chen2024drct} & 85.18 & 71.96 & 85.20 & 79.80 & 81.76 & 67.17 & 84.05 & 72.98 \\
DGS-Net~\citep{yan2026dgsnet} & 87.80 & \underline{82.30} & 42.90 & 59.41 & 49.97 & 51.13 & 60.22 & 64.28 \\
HiDA-Net~\citep{mu2026hida} & N/A & N/A & N/A & 79.10 & N/A & 80.33 & N/A & N/A \\
DDA~\citep{chen2025dda} & \underline{90.20} & 81.60 & \underline{91.20} & \underline{82.40} & \underline{93.70} & \underline{85.09} & \underline{91.70} & \underline{83.03} \\
\midrule
\leavevmode\rlap{\textcolor{oursLav}{\rule[-\arraystretch\dp\strutbox]{\textwidth}{\dimexpr\arraystretch\ht\strutbox+\arraystretch\dp\strutbox\relax}}}\textbf{RGE (Ours)} & \textbf{97.58} & \textbf{93.41} & \textbf{98.33} & \textbf{94.74} & \textbf{98.34} & \textbf{93.50} & \textbf{98.08} & \textbf{93.88} \\
\bottomrule
\end{tabularx}
\endgroup
\end{minipage}
\end{table*}

\subsection{Targeted Adaptation}
\label{sec:reserve-elicitation}

\noindent\textbf{Reference Detector.}
We construct a reference detector from the frozen backbone's pooled representations.
For $N$ discovery fitting images, we assign weights $\omega_i=1/(2N_{y_i})$, where $N_y$ is the number of images in class $y$, so that $\sum_i\omega_i=1$.
We standardize the representations using their weighted empirical means and population standard deviations, then fit a weighted $\ell_2$-regularized logistic classifier with $C=1$.
The standardization statistics remain fixed throughout adaptation and inference.
A logit threshold $\tau$ is selected to maximize balanced accuracy on the held-out development set.
For a standardized pooled representation $v$, the scoring head is $\psi(v)=w^\top v+b-\tau$.
It stays fixed during FRA training, with $\psi=0$ as the decision boundary.

\noindent\textbf{Adapter Placement.}
FRA inserts a residual branch at each selected site $c\in\mathcal C_\star$.
Each branch operates token-wise within its designated group, sharing parameters across tokens at that site while using separate parameters at different sites.
Selected patch branches operate on all native patch tokens.
A branch reads the current hidden state $h_{c,p}$, which includes the effects of writes in earlier blocks, and passes the updated state to subsequent computation.
Let $\Phi_{\theta,\mathcal A}(x)$ denote the pooled backbone output after the fixed standardization, with pretrained parameters $\theta$ and coefficient maps $\mathcal A=\{A_c:c\in\mathcal C_\star\}$.
The adapted detector score is
\begin{equation}
f_{\mathcal A}(x)=\psi\!\left(\Phi_{\theta,\mathcal A}(x)\right).
\label{eq:detail-detector-score}
\end{equation}

\noindent\textbf{Residual Update.}
At each selected site, FRA updates the current hidden state as
\begin{equation}
h'_{c,p}=h_{c,p}+U_cA_ch_{c,p},
\label{eq:fra-update}
\end{equation}
where $A_c\in\mathbb R^{r_c\times d_c}$ is the learned coefficient map.
The map $A_c$ reads the complete current hidden state to produce input-dependent coefficients, and the fixed basis $U_c$ maps these coefficients into the reserve subspace.
Each token therefore receives its own residual write within $\operatorname{span}(U_c)$.
The residual map $U_cA_c$ has rank at most $r_c$, with its column space contained in the mapped reserve subspace.
Whereas standard low-rank adaptation (LoRA) learns both factors of an update, FRA fixes its output basis through reserve localization and learns only the coefficient map.

\noindent\textbf{Optimization.}
Each $A_c$ is initialized to zero, so the augmented model initially reproduces the reference detector.
We optimize only the coefficient maps on the discovery fitting images using weighted binary cross-entropy (BCE), with $f_{\mathcal A}(x)$ as the logit:
\begin{equation}
\mathcal L(\mathcal A)
=\sum_{i=1}^{N}\omega_i\,
\ell_{\mathrm{BCE}}\!\left(f_{\mathcal A}(x_i),y_i\right).
\label{eq:detail-detection-loss}
\end{equation}
Gradients propagate through the frozen backbone and scoring head to update $\mathcal A$, while the backbone parameters, classifier, and writing bases remain fixed.
Detection supervision thus learns how to combine the selected writing directions for each input.
The trainable parameter count is $\sum_{c\in\mathcal C_\star}r_cd_c$, determined by the selected sites and their writing-subspace dimensions.
At inference, the augmented backbone and retained scoring head directly score query images using the same fixed standardization.
The appendix details basis construction and optimization.

\newlength{\savedTextGap}\setlength{\savedTextGap}{\textfloatsep}
\newlength{\savedInGap}\setlength{\savedInGap}{\intextsep}
\newlength{\savedFloatGap}\setlength{\savedFloatGap}{\floatsep}
\newlength{\savedDoubleTextGap}\setlength{\savedDoubleTextGap}{\dbltextfloatsep}
\newlength{\savedDoubleFloatGap}\setlength{\savedDoubleFloatGap}{\dblfloatsep}
\setlength{\textfloatsep}{8pt plus 1pt minus 1pt}
\setlength{\floatsep}{6pt plus 1pt minus 1pt}
\setlength{\dbltextfloatsep}{8pt plus 1pt minus 1pt}
\setlength{\dblfloatsep}{6pt plus 1pt minus 1pt}
\begin{table}[!t]
\begingroup

\centering
\caption{\textbf{Adaptation Structure on AIGIBench.} Parentheses show changes in mAP and mBA relative to RGE.}

\label{tab:structure-allocation}
\vspace{-0.3cm}
\begingroup
\fontsize{8.2}{9.5}\selectfont
\setlength{\tabcolsep}{0pt}
\renewcommand{\arraystretch}{1.0}
\newcommand{\dlt}[1]{{\fontsize{6}{7}\selectfont\,\textcolor{deltaRed}{(#1)}}}
\begin{tabularx}{\linewidth}{@{}>{\raggedright\arraybackslash}X l l >{\hspace{6pt}}l l@{}}
\toprule
\textbf{Adaptation Variant} & \multicolumn{2}{l}{\textbf{mAP} $\uparrow$} & \multicolumn{2}{>{\hspace{6pt}}l@{}}{\textbf{mBA} $\uparrow$} \\
\midrule
\multicolumn{5}{@{}l}{\textcolor{black!55}{(a)~Component Selection}} \\
\noalign{\vskip\arrayrulewidth\hbox to\linewidth{\leaders\hbox{\vrule width1.5pt height\arrayrulewidth\hskip1.5pt}\hfill}\vskip\arrayrulewidth}\noalign{\vskip 1pt}
Random-In & 95.37 & \dlt{\textminus 2.21} & 89.52 & \dlt{\textminus 3.89} \\
Random-Ex & 94.26 & \dlt{\textminus 3.32} & 88.47 & \dlt{\textminus 4.94} \\
Bottom-K & 93.31 & \dlt{\textminus 4.27} & 87.34 & \dlt{\textminus 6.07} \\
\addlinespace[3pt]
\multicolumn{5}{@{}l}{\textcolor{black!55}{(b)~Allocation and Parameterization}} \\
\noalign{\vskip\arrayrulewidth\hbox to\linewidth{\leaders\hbox{\vrule width1.5pt height\arrayrulewidth\hskip1.5pt}\hfill}\vskip\arrayrulewidth}\noalign{\vskip 1pt}
CLS-Budget Reallocation & 93.01 & \dlt{\textminus 4.57} & 89.50 & \dlt{\textminus 3.91} \\
Free-LoRA & 95.14 & \dlt{\textminus 2.44} & 88.79 & \dlt{\textminus 4.62} \\
Random Weight Coordinates & 93.08 & \dlt{\textminus 4.50} & 86.13 & \dlt{\textminus 7.28} \\
\midrule
Baseline & 91.90 & \dlt{\textminus 5.68} & 68.86 & \dlt{\textminus 24.55} \\
\rowcolor{oursLav}\textbf{RGE (Ours)} & \textbf{97.58} & & \textbf{93.41} & \\
\bottomrule

\end{tabularx}
\endgroup
\endgroup
\end{table}

\section{Experiments and Analysis}

\label{sec:experiments}

\noindent\textbf{Discovery Set and Benchmarks.}
We perform localization and adaptation on a labeled discovery set of real and fake images; the appendix details its construction from public sources. Our main comparison evaluates AIGIBench~\citep{li2025aigibench}, Chameleon~\citep{yan2025aide}, and HiRes-50K~\citep{mu2026hida}, covering diverse generation mechanisms, perceptually convincing images, and high-resolution content. These benchmarks supply no training images for the classifier or FRA. Additional evaluations and source breakdowns appear in the appendix.

\noindent\textbf{Evaluation Metrics.} 
We report mean average precision (mAP) for ranking quality and follow each benchmark's native accuracy (Acc) convention in Table~\ref{tab:main-results}; on AIGIBench, Acc is the mean balanced accuracy (mBA) over real and fake images, which the AIGIBench analyses also report. Evaluation and metric aggregation follow the official protocol specified for each benchmark.

\noindent\textbf{Implementation.} 
DINOv3-H+~\citep{simeoni2025dinov3} serves as our primary backbone. We use 5,000 labeled discovery images by default. RGE retains the fitted pooled-feature classifier and learns only FRA coefficient maps, with the backbone and writing bases fixed. 
Zero-initialized coefficient maps preserve the reference detector's predictions at the start of adaptation.
We select the top 1\% of candidate components by origin sensitivity. The resulting writing ranks vary with discovery size; the 0.5K, 5K, and 20K configurations each use a trainable parameter budget below 0.2\% of the backbone (the appendix). 
Training configurations appear in the appendix and the component-budget analysis in Section~\ref{sec:component-budget}.

\setlength{\textfloatsep}{7pt plus 1pt minus 1pt}
\setlength{\intextsep}{7pt plus 1pt minus 1pt}
\setlength{\floatsep}{7pt plus 1pt minus 1pt}

\subsection{Comparison with State-of-the-Art Methods}

\label{sec:performance}

\noindent\textbf{Quantitative Results.} 
We first compare RGE with eight state-of-the-art detectors across AIGIBench~\citep{li2025aigibench}, Chameleon~\citep{yan2025aide}, and HiRes-50K~\citep{mu2026hida} without target-benchmark adaptation (Table~\ref{tab:main-results}). Representative frozen-readout and artifact-based detectors such as UnivFD~\citep{ojha2023univfd} and AIDE~\citep{yan2025aide} remain limited on Chameleon and HiRes-50K, reaching 56.46\%--65.80\% Acc. DGS-Net~\citep{yan2026dgsnet}, adapted on the 144K-image AIGIBench split, attains the second-best Acc on AIGIBench (82.30\%) but drops to 59.41\% and 51.13\% Acc on Chameleon and HiRes-50K. In contrast, RGE achieves the highest mAP and Acc on all three benchmarks, averaging 98.08\% mAP and 93.88\% Acc and surpassing the strongest overall competitor, DDA~\citep{chen2025dda}, by 6.38 mAP and 10.85 Acc points. Notably, RGE exceeds DDA by 7.13 mAP and 12.34 Acc points on Chameleon, while on HiRes-50K it outperforms DDA by 4.64 mAP and 8.41 Acc points and the detail-preserving HiDA-Net~\citep{mu2026hida} by 13.17 Acc points. Whereas the strongest competitors train on 144K--473K images (the appendix), RGE achieves these gains using 5,000 discovery images, eliciting forensic evidence already encoded in the pretrained backbone; source- and resolution-level breakdowns appear in the appendix.

\begin{figure}[!t]
\centering
\includegraphics[width=0.98\linewidth]{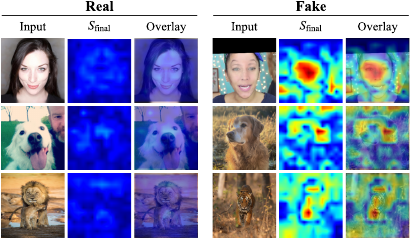}
\vspace{-0.3cm}
\caption{\textbf{Final-Score Sensitivity.} Final-score sensitivity $S_{\mathrm{final}}$ for real and fake image pairs (blue and red indicate low and high sensitivity, respectively).}
\label{fig:fra-spatial}
\end{figure}

\noindent\textbf{Visualization.}
Figure~\ref{fig:fra-spatial} visualizes the final-score sensitivity $S_{\mathrm{final}}$ across independent real and fake portrait, dog, and large-cat pairs to examine which image regions drive RGE's decision. In each fake image, sensitivity concentrates sharply on the synthesized subject---such as the portrait face, the dog's head, and the tiger's face and foreleg---whereas real images show no such peaks. Spatial sensitivity isolated for the elicited writes peaks on the same subject regions (see the appendix for localized-write maps and perturbation details), showing that the reserve-elicited writes contribute to the final real--fake prediction by selectively responding to the synthesized subject.

\subsection{Ablation Studies}

\label{sec:ablation}

\begin{figure}[!t]
\centering
\includegraphics[width=0.98\linewidth]{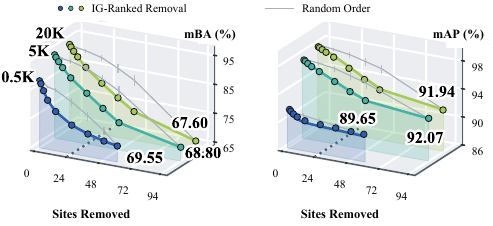}
\vspace{-0.3cm}
\caption{\textbf{Cumulative Removal of Learned Writes.} Removing writes in gate-IG order versus five random orders (mean and range): mBA (left) and mAP (right).}
\label{fig:fra-cumulative-ig}

\par\vspace{10pt}
\centering
\includegraphics[width=0.98\linewidth]{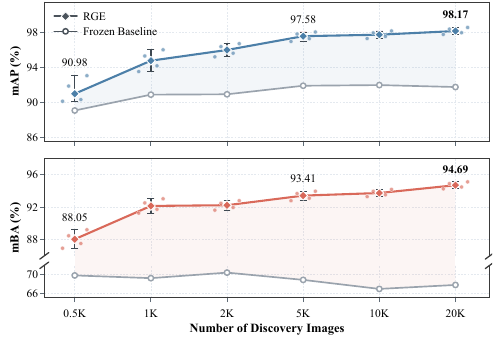}
\vspace{-0.3cm}
\caption{\textbf{Scaling with Discovery Size.} Detection across discovery sizes on AIGIBench; bars span five training seeds.}
\label{fig:fra-scaling}

\end{figure}

\noindent\textbf{Adaptation Structure.}
\label{sec:adaptation-structure}
To isolate what the reserve structure contributes, we alter its components, their allocation, or the constraint on the update. Random-In keeps RGE's sites and per-site counts but draws each site's components at random from its non-reserve ICA components, while Bottom-K retains the components with the lowest origin sensitivity at the same sites and counts. Random-Ex excludes all reserve components and draws the same total number of components across the full grid, with sites determined by the draw. CLS-Budget Reallocation moves the CLS budget to patch/register sites. Free-LoRA learns unconstrained low-rank updates to the attention-output and MLP-down projections of the layers that contain reserve sites, with ranks set to match RGE's trainable-parameter count. Random Weight Coordinates learns updates to scalar entries sampled uniformly without replacement from all backbone linear weight matrices, with the number of selected entries matched to RGE's trainable-parameter count. All adapted variants use matched trainable capacity. The three selection controls form a graded comparison (Table~\ref{tab:structure-allocation}~(a)): random non-reserve components at RGE's sites (Random-In) cost less than random components across the full grid (Random-Ex), and the lowest-scoring components at the same sites (Bottom-K) cost the most. Detection therefore follows the F-lens score both within and across sites, and components outside the reserve do not fully substitute for it. At matched capacity, reallocating the CLS budget or removing the subspace constraint also falls short of RGE (Table~\ref{tab:structure-allocation}~(b)), with Free-LoRA trailing by 2.44 mAP; these comparisons link the gain to where residual updates enter the backbone and which feature directions they follow.

\begingroup\clubpenalties 4 10000 10000 10000 0\relax
\noindent\textbf{Cumulative Removal.}
Mean label-aligned gate integrated gradients (IG)~\citep{sundararajan2017axiomatic} attribute score changes to learned writes. Let $f(x;\mathbf g)$ denote the score when a gate $g_c\in[0,1]$ scales the write at each site $c$ in Eq.~(\ref{eq:fra-update}), with learned parameters fixed, so that $\mathbf g=\mathbf0$ recovers the reference detector and $\mathbf g=\mathbf1$ gives RGE. Integrating along $\mathbf g=t\mathbf1$ gives
\begin{equation}
\mathrm{IG}_c(x)=\int_0^1\frac{\partial f(x;t\mathbf 1)}{\partial g_c}\,dt.
\label{eq:main-gate-ig}
\end{equation}
By completeness, $\sum_c\mathrm{IG}_c(x)=f(x;\mathbf1)-f(x;\mathbf0)$, partitioning the score change from the reference detector to RGE across sites. Averaging $(2y-1)\mathrm{IG}_c(x)$ over analysis images aligns contributions with the correct class (the appendix). Having examined how discovery structures learning, we test how the learned FRA writes sustain detection. Figure~\ref{fig:fra-cumulative-ig} progressively disables writes in the trained 0.5K, 5K, and 20K detectors, following either their gate-IG ranking, computed on a fixed, stratified AIGIBench analysis panel (see appendix) for each detector, or random orders. By 24 disabled sites, IG-ranked removal yields lower mAP and mBA than the random-order mean in all three configurations. Removing all writes reduces mAP by 1.69--6.01 points and mBA by 17.65--25.50 points. These interventions leave the backbone and classifier fixed and involve no retraining, directly connecting the learned internal updates to ranking quality and balanced classification across all three discovery sizes.

\begin{figure}[!t]
\centering
\includegraphics[width=0.98\linewidth]{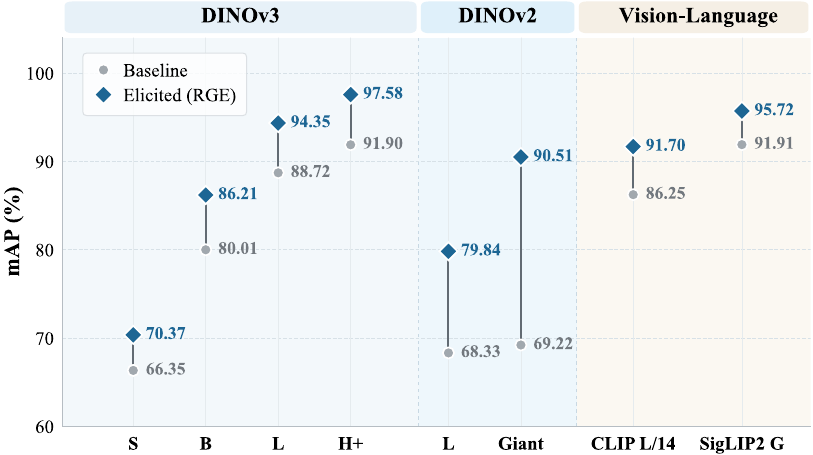}
\vspace{-0.3cm}
\caption{\textbf{Cross-Backbone Evaluation.} mAP of the baseline (gray) and RGE (blue) across eight ViT backbones on AIGIBench.}
\label{fig:fra-backbones}

\end{figure}

\noindent\textbf{Scaling with Discovery Data.}
Figure~\ref{fig:fra-scaling} examines how RGE scales with discovery data to reveal the full potential of the forensic reserve in VFMs. With only 500 discovery images, RGE already reaches 90.98\% mAP and 88.05\% mBA on AIGIBench, surpassing the competing methods in Table~\ref{tab:main-results}. RGE improves over the baseline across all discovery sizes, reaching 97.58\% mAP and 93.41\% mBA at 5K, and 98.17\% mAP and 94.69\% mBA at 20K. 
The 0.5K pool therefore does not exhaust the reserve: larger discovery pools continue to strengthen its observable expression. Gains taper beyond 5K, where quadrupling the pool to 20K adds 0.59 mAP and 1.28 mBA points. This pattern suggests that a compact discovery set suffices to localize and elicit much of the accessible forensic reserve, while broader coverage helps refine component selection and input-dependent writes. The 5K configuration balances elicitation of the pretrained model's forensic capacity with labeled data requirements, motivating its use as our default setting.

\begin{figure}[!t]

\centering
\includegraphics[width=0.98\linewidth]{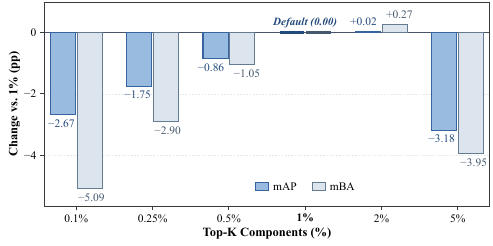}
\vspace{-0.3cm}
\caption{\textbf{Component Selection Budget.} Change in mAP and mBA relative to the 1\% budget on AIGIBench.}
\label{fig:fra-budget}
\end{figure}

\begingroup\clubpenalty=10000
\begin{figure}[!t]
\centering
\includegraphics[width=0.98\linewidth]{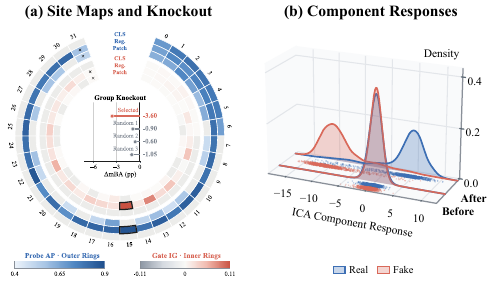}
\vspace{-0.3cm}
\caption{\textbf{Reserve Localization and Elicited Responses.} (a) Probe average precision (AP; outer rings) and gate IG (inner rings) per layer--token site, with the knockout of discovery-ranked versus random site groups at the center. (b) Responses of one reserve component on the same images before and after FRA, measured in a fixed ICA coordinate.}
\label{fig:fra-functional-writes}\label{fig:component-separation}

\end{figure}

\subsection{Further Analysis}

\label{sec:vision-models}

\noindent\textbf{Cross-Backbone Evaluation.}
Reserve-Guided Elicitation extends across ViT architectures, capacities, and pretraining families. Figure~\ref{fig:fra-backbones} compares eight backbones from DINOv3~\citep{simeoni2025dinov3}, DINOv2~\citep{oquab2023dinov2}, CLIP~\citep{radford2021clip}, and SigLIP2~\citep{tschannen2025siglip2}, each with its fitted classifier. RGE improves mAP over every corresponding baseline. DINOv3-H+ reaches 97.58\% mAP, while DINOv2-L and DINOv2-Giant gain 11.51 and 21.29 points, respectively. Enhancement therefore benefits both a strong initial ranking representation and models with larger room for improvement. Each pretraining family carries its own forensic repertoire, yet the same localize-and-elicit procedure recruits a reserve from it in every case, with the attainable improvement depending on the model rather than on family-specific tuning of the procedure.

\begingroup\clubpenalties 4 10000 10000 10000 0\relax
\noindent\textbf{Component Selection Budget.}
\label{sec:component-budget}
The retained component fraction also affects detection. In the budget sweep, increasing the fraction from 0.1\% to 1\% improves both metrics; 2\% adds only 0.02 mAP and 0.27 mBA points, while 5\% reduces both relative to 1\%.
A compact reserve therefore suffices: beyond about 1\% of the candidate components, additional components contribute little and eventually reduce detection. The default 1\% lies near the best observed performance (Figure~\ref{fig:fra-budget}) while keeping trainable parameters below 0.2\% of the backbone: selected components span most sites with small per-site ranks, concentrating capacity on the strongest directions within each site (the appendix reports writing ranks and trainable capacities).
\par\endgroup

\noindent\textbf{Reserve Localization.}
\label{sec:functional-writes}
Forensic information is distributed unevenly across the layer--token sites analyzed by F-lens. Figure~\ref{fig:fra-functional-writes}~(a) relates this internal organization to the learned writes. Out-of-fold probe AP, shown on the outer rings, measures pre-adaptation separability on the discovery set using token-averaged component responses and five-fold logistic probes. The inner rings show gate IG. Both maps are uneven, and the layer-15 register site is strongest in both: a location with highly accessible forensic information also makes a prominent contribution after adaptation. The reserve is unevenly weighted across sites, and the sites where it is most accessible before adaptation are among those where elicitation contributes most. Prominent responses occur across CLS, register, and patch tokens. Together with the performance loss from CLS-Budget Reallocation in Table~\ref{tab:structure-allocation}~(b), this distribution supports reserve-guided allocation of adaptation capacity across token groups.

\begin{figure}[!t]
\centering
\includegraphics[width=0.98\linewidth]{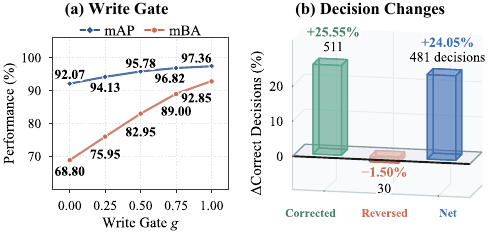}
\vspace{-0.3cm}
\caption{\textbf{Write Gate and Decision Changes.} (a) Detection versus write gate $g$. (b) Decisions changed by the writes.}
\label{fig:writing-gate}
\end{figure}

\noindent\textbf{Site Knockout.}
To test by intervention whether this discovery-set ranking transfers to held-out images, we disable the six writing sites with the highest probe AP and compare them with three random six-site groups of approximately matched total writing rank. On the AIGIBench analysis panel, removing the discovery-selected group reduces mBA by 3.60 points, versus 0.60--1.05 points for the random groups (Figure~\ref{fig:fra-functional-writes}~(a), center). Measured without retraining the detector, this mBA loss is 3.4--6.0 times that of a random group with approximately matched writing rank, so the ranking that discovery fixes before adaptation transfers to held-out benchmark images and identifies the sites where the elicited writes matter most for detection.
\par\endgroup

\noindent\textbf{Component-Level Separation.}
\label{sec:learned-writes}
Within a single reserve component, we observe elicitation by comparing its responses in a fixed ICA coordinate before and after adaptation.
Figure~\ref{fig:component-separation}~(b) shows substantially reduced overlap between real and fake image responses on the same images.
After adaptation, real-image responses shift toward higher coordinate values, while fake-image responses shift toward lower values.
With the discovery-fitted coordinate and readout held fixed, the observed separation reflects changes in the component response. The writing basis $U_c$ is fixed, but the coefficients $A_c h_{c,p}$ in Eq.~(\ref{eq:fra-update}) depend on each input's current hidden state. These input-dependent updates provide a mechanism for reinforcing distinct real and fake responses along a shared reserve direction.

\begin{table}[!t]

\begingroup
\captionsetup{position=top}
\centering
\caption{\textbf{Depth-Restricted Adaptation.} Detection at the same total component budget. Depth-restricted variants select components within one band; RGE selects across all depths.}
\label{tab:writing-depth}
\vspace{-0.3cm}
\begin{tabularx}{\linewidth}{@{}>{\raggedright\arraybackslash}Xcc@{}}
\toprule
\textbf{Depth Band} & \textbf{mAP} $\uparrow$ & \textbf{mBA} $\uparrow$ \\
\midrule
Shallow & 94.25 & 88.56 \\
Middle & 97.21 & 92.92 \\
Deep & 96.37 & 91.71 \\
\midrule
\leavevmode\rlap{\textcolor{oursLav}{\rule[-\arraystretch\dp\strutbox]{\columnwidth}{\dimexpr\arraystretch\ht\strutbox+\arraystretch\dp\strutbox\relax}}}\textbf{RGE (Ours)} & \textbf{97.58} & \textbf{93.41} \\
\bottomrule
\end{tabularx}
\endgroup
\end{table}

\begingroup\clubpenalties 4 10000 10000 10000 0\relax
\noindent\textbf{Write Gate.}
Varying the write gate reveals how the trained residual updates translate into individual decisions. On the AIGIBench analysis panel, we increase a global write gate $g$ (defined in the appendix) from zero to one without retraining, while retaining the classifier's decision boundary. mBA rises from 68.80\% to 92.85\% across the tested gate levels (Figure~\ref{fig:writing-gate}~(a)). Restoring the learned writes corrects 511 previously incorrect predictions and reverses 30 correct ones, yielding 481 additional correct decisions (\mbox{Figure~\ref{fig:writing-gate}~(b)}).
Alongside rising mAP (from 92.07\% to 97.36\%), the elicited writes change thresholded decisions overwhelmingly in the correct direction, with 17 corrections for every reversal of an initially correct prediction. Together with the component-level separation in Figure~\ref{fig:component-separation}~(b), these results link enhanced internal responses to improved ranking and decision accuracy under the frozen readout.
\par\endgroup

\noindent\textbf{Writing Depth.}
We next restrict component selection and adaptation to one depth band at a time, keeping the total component budget fixed, to examine how depth affects elicitation (Table~\ref{tab:writing-depth}). The shallow, middle, and deep bands cover backbone layers L0--10, L11--21, and L22--31, respectively. Middle-layer writes achieve 97.21\% mAP and 92.92\% mBA, recovering 93.5\% and 98.0\% of full RGE's respective improvements over the frozen baseline. Both scores exceed those of the shallow- and deep-layer variants, while full RGE performs best at 97.58\% mAP and 93.41\% mBA. 
Most of the gain is therefore recoverable within the middle band, and the global budget of full RGE, which also retains shallow and deep components, adds the remaining improvement over the middle band alone.

\noindent\textbf{Robustness to Image Perturbations.}
The benefit of the learned writes also persists under common image perturbations. Figure~\ref{fig:fra-robustness-gain}~(a) evaluates clean images and six independently applied JPEG, resizing, and blurring conditions on the AIGIBench analysis panel. RGE maintains a 3.03--10.64-point mAP advantage over the baseline across all seven conditions. Resizing preserves its mAP, JPEG compression widens the advantage, and blurring narrows it. Even blurring, which removes the most fine image detail, leaves the elicited detector more discriminative than its baseline, so the elicited evidence is not confined to fragile high-frequency traces. Figure~\ref{fig:fra-robustness-gain}~(b) provides example spatial responses under these perturbations. Across all six perturbations, the high-sensitivity region in the fake example remains at the same spatial location, while final-score sensitivity in the real example remains uniformly low throughout the image across all tested conditions.
\begin{figure}[!t]
\centering
\begin{minipage}[t]{0.435\linewidth}\vspace{0pt}\centering{\fontsize{8}{9}\selectfont\bfseries (a) mAP Under Perturbations}\par\vspace{2pt}
\includegraphics[width=0.98\linewidth]{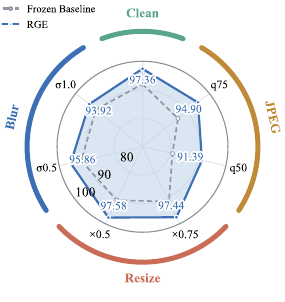}\end{minipage}\hfill
\begin{minipage}[t]{0.535\linewidth}\vspace{0pt}\centering{\fontsize{8}{9}\selectfont\bfseries (b) Sensitivity Under Perturbations}\par\vspace{2pt}
\includegraphics[width=0.98\linewidth]{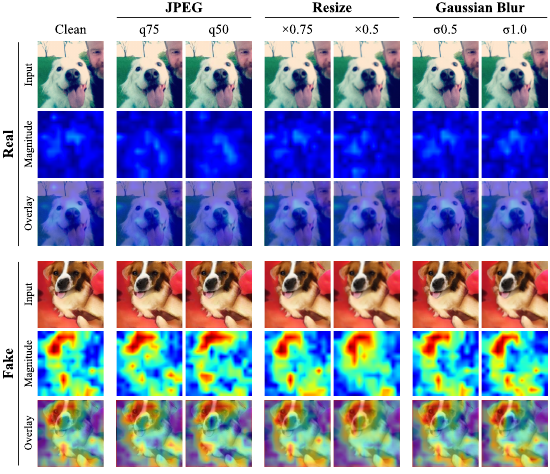}\end{minipage}
\vspace{-0.3cm}
\caption{\textbf{Robustness Under Image Perturbations.} (a) mAP of RGE and the baseline on clean images and under JPEG compression, resizing, and blurring. (b) Final-score sensitivity for a real (top) and a fake (bottom) dog under the same perturbations.}
\label{fig:fra-robustness-gain}\label{fig:fra-robustness-spatial}
\end{figure}

\setlength{\textfloatsep}{\savedTextGap}
\setlength{\intextsep}{\savedInGap}
\setlength{\floatsep}{\savedFloatGap}
\setlength{\dbltextfloatsep}{\savedDoubleTextGap}
\setlength{\dblfloatsep}{\savedDoubleFloatGap}

\begingroup\clubpenalties 4 10000 10000 10000 0\relax
\section{Conclusion}
\label{sec:conclusion}

We introduced RGE to discover forensic reserve within vision models and turn it into targeted detection enhancement. F-lens localizes origin-sensitive components whose locations specify adaptation sites and whose mapped directions define fixed writing subspaces; the Forensic Reserve Adapter learns input-dependent residual writes within them. Our findings show that internal forensic knowledge supplies both evidence for classification and concrete targets for enhancing detection: the sites that discovery ranks highest before adaptation are where the elicited writes matter most, matched-capacity controls confirm the role of reserve selection and subspace constraints, and the learned writes correct 511 previously incorrect decisions while reversing only 30. With 5,000 labeled training images, under 0.2\% trainable parameters, and no target-benchmark adaptation, RGE achieves the best mAP and Acc among the compared methods on all three main benchmarks; its gains also extend across eight ViT backbones and persist under common image perturbations. More broadly, pretrained vision models hold a forensic repertoire that pooled readout leaves underused; locating a reserve within it and eliciting that reserve offers a data-efficient route to detection that does not depend on ever larger labeled forgery corpora.
\par\endgroup

\subsection*{LLM Usage Statement}
We used OpenAI Codex to assist with manuscript editing, reference checking, and LaTeX and figure preparation. The authors reviewed the AI-assisted content and take full responsibility for the paper.

\clearpage

\setcounter{page}{1}
\renewcommand{\thepage}{S\arabic{page}}
\twocolumn[\begin{center}
{\Large\bfseries\thetitle\par}
\medskip
{\Large Supplementary Material\par}
\end{center}]
\appendix
\input{generator-citations}

\setlength{\intextsep}{6pt}
\captionsetup{skip=3pt}
\captionsetup[table]{position=top}
\setlength{\textfloatsep}{8pt plus 1pt minus 1pt}
\setlength{\dbltextfloatsep}{8pt plus 1pt minus 1pt}
\setlength{\floatsep}{6pt plus 1pt minus 1pt}
\setlength{\dblfloatsep}{8pt plus 1pt minus 1pt}
\clubpenalty=10000 \widowpenalty=10000
\makeatletter
\setlength{\@fptop}{0pt}\setlength{\@dblfptop}{0pt}
\setlength{\@fpsep}{18pt}
\setlength{\@fpbot}{0pt plus 1fil}
\makeatother
\setcounter{topnumber}{4}
\setcounter{totalnumber}{4}
\raggedbottom
\renewcommand{\dblfloatpagefraction}{.72}\renewcommand{\floatpagefraction}{.72}\numberwithin{equation}{section}
\setlength{\parskip}{0pt}
\section*{Appendix}
We cover discovery-set and baseline protocols (Section~\ref{app:experimental-settings}), F-lens and FRA implementation and optimization (Section~\ref{app:detailed-methodology}), functional and capacity analyses (Section~\ref{app:additional-analyses}), results by source and resolution, including Treasure-64 (Section~\ref{app:additional-results}), and limitations and future work (Section~\ref{app:limitations}). The appendix closes with a notation guide (Table~\ref{tab:reserve-notation}).

\section{Experimental Settings}
\label{app:experimental-settings}
\noindent\textbf{Discovery Set Construction.}
We construct class-balanced discovery pools from public datasets independently of all evaluation benchmarks, balancing coverage across sources and limiting repeated frames from individual videos so that no single generator or capture format dominates localization. Fake images cover GAN synthesis and pixel-space diffusion~\citep{park2025communityforensics}, latent diffusion~\citep{chen2026aligngemini,livernoche2025openfake,openfakev2data}, commercial generation and editing~\citep{zhang2026benchmark,zhou2026pgc}, and face swapping, editing, and synthesis~\citep{shao2023dgm4,yan2024df40}. Real images cover natural and social photography~\citep{yang2016widerface,unsplashlite,kuznetsova2020openimages}, human-designed graphics~\citep{chen2025postercraft}, documents and receipts~\citep{pfitzmann2022doclaynet,park2019cord}, and archival scans~\citep{locfreetouse,bl2014bookimages}. At each discovery budget, we report RGE's gain over the frozen reference detector fitted on the same images.

\noindent\textbf{Quality Control and Deduplication.}
Each image record tracks provenance, content domain or generator family, native resolution, codec, and EXIF-oriented RGB8 pixel and 64-bit perceptual hashes. We retain decodable primary images with verified labels (including any official preprocessing) and exclude corrupted files, auxiliary assets, and text-only edits. For video sources, we keep at most four evenly spaced frames per generator and source video to limit temporal repetition. We then deduplicate within the real and fake classes separately, since face swaps and local edits often closely resemble their real originals. Within each class, we remove exact repeats by pixel and source identity, and merge near-duplicates with a 64-bit perceptual-hash distance of at most two bits and an aspect-ratio difference of at most 1\% (compared across sources for real images and within each generator for fake images). Representatives favor larger native images and then lossless encoding, with seeded tie-breaking; pixel-identical records with conflicting labels are handled separately as label conflicts.

\begin{figure*}[t]
\centering
\includegraphics[width=0.98\linewidth]{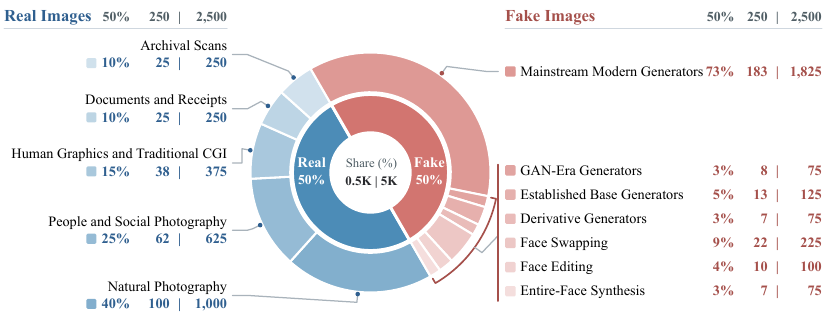}

\caption{\textbf{Discovery Composition.} Wedges show within-class target shares. Labels give each share and the counts of unique images selected for the 0.5K and 5K pools, respectively, before training draws with replacement.}
\label{fig:discovery-sunburst}
\end{figure*}
\noindent\textbf{Group-Disjoint Partitioning and Hierarchical Sampling.}
Each reported discovery budget counts the fitting images and pairs that fitting pool with an additional held-out development set equal to 20\% of the fitting pool size (100 development images at 0.5K). We split the data by complete source-origin groups so that an original image, its edited derivatives, and its near-duplicates always stay on the same side of the split. We use the development set to select the reference threshold $\tau$ and for forward evaluation with the discovered sites and F-space held fixed; its images are excluded from component discovery, classifier fitting, and adapter optimization. Each fitting pool is balanced equally between real and fake images. Within each class, we first allocate quotas across content domains or generation categories according to target weights (Figure~\ref{fig:discovery-sunburst}) using largest-remainder rounding. Within each real domain, quotas are divided equally across underlying datasets before cycling through dataset variants. Within each fake category, after subdivision by capture format or source type where applicable, quotas are divided equally across generator families and then rotated across individual generators with a fixed seed. If a group runs out of eligible images, we redistribute its remaining quota among groups in the same category. The category therefore retains its target count even when an individual group cannot fill its assigned quota.

\noindent\textbf{Generator Caps and Resolution Matching.}
To prevent individual sources or formats from dominating the fake pool, we cap any single generator at 4\%, cropped face datasets at 12\%, and video frames at 16\% of all fake images. Derivative fine-tuned generators are capped at 3\%, and GAN-era, established base, and derivative generators together are capped at 12\%. We also match resolution distributions across real and fake images using four longest-edge bands: below 640, 640--1,023, 1,024--2,047, and at least 2,048 pixels. For pools up to 5K, the target shares across these four bands are 28\%, 5\%, 57\%, and 10\% within each class (with a $\pm 10$ percentage-point tolerance per band); for 10K and 20K, the highest-resolution fake band uses a 3\% lower bound instead. At every scale, real and fake shares within any band differ by at most 14 percentage points. When a band exceeds its target, excess images are replaced first from the same dataset or generator, and otherwise from the same domain or generator family.

\noindent\textbf{Nested Discovery Budgets.}
The six discovery pools (0.5K, 1K, 2K, 5K, 10K, and 20K) are nested: each larger pool contains all images from the smaller pools and applies quota or resolution replacements only to newly added images. Figure~\ref{fig:discovery-sunburst} shows the target category weights and realized counts at 0.5K and 5K. Within the mainstream modern generator category, native full-frame images and downsampled rasters follow an 85/15 split; within face swapping, full news images and 256-pixel face crops are balanced 50/50.

\noindent\textbf{Training Configuration.}
We train FRA using AdamW with learning rate $3\times 10^{-4}$, zero weight decay, and gradient-norm clipping at $1.0$. Each epoch draws $N$ images uniformly with replacement from the balanced fitting pool. The batch size is 64 for the 0.5K and 1K pools and 256 for the 2K--20K pools. We train for four epochs at 0.5K--5K, two at 10K, and one at 20K, so the 5K, 10K, and 20K detectors all train for the same total budget of 20,000 image draws. Feature standardization, reference-classifier fitting, and the training loss follow Section~\ref{sec:reserve-elicitation} and Appendix~\ref{app:detailed-methodology}.

\noindent\textbf{Benchmark Baselines and Evaluation Protocol.}
Table~\ref{tab:main-results} compares RGE (5,000 discovery images) with UnivFD~\citep{ojha2023univfd}, DRCT~\citep{chen2024drct}, AIDE~\citep{yan2025aide}, DDA~\citep{chen2025dda}, HiDA-Net~\citep{mu2026hida}, DGS-Net~\citep{yan2026dgsnet}, LTD~\citep{yang2026ltd}, and PLM~\citep{zhou2026plm} (counts below refer to detector training, excluding backbone pretraining):
\begin{itemize}
    \item \emph{AIGIBench}: DGS-Net trains CLIP LoRA adapters with distillation-guided gradient surgery on the 144K-image Setting-II split~\citep{li2025aigibench,yan2026dgsnet}. DDA trains DINOv2 LoRA adapters on 236K images (118K COCO originals and their 118K SD2.1 VAE reconstructions, aligned in the pixel and frequency domains)~\citep{chen2025dda,li2025aigibench}.
    \item \emph{Chameleon}: AIDE trains a DCT-patch and CLIP-ConvNeXt detector on full GenImage~\citep{yan2025aide}. DRCT uses the 473K-image SD2.1 setup reported by GAPL~\citep{chen2024drct,qin2026gapl}, and DDA uses the 236K setup above~\citep{chen2025dda}.
    \item \emph{HiRes-50K}: Reported Acc scores for UnivFD, AIDE, DRCT (ConvB), and HiDA-Net come from the benchmark paper's full-GenImage evaluation, macro-averaged across its eight resolution intervals~\citep{mu2026hida}.
\end{itemize}
Where available, we directly adopt the values reported in official benchmark comparisons. For missing cells where official checkpoints are available, we evaluate the released weights under each benchmark's standard protocol: DRCT (mAP and Acc) on AIGIBench; DDA (mAP) and DGS-Net, LTD, and PLM (mAP and Acc) on Chameleon; and UnivFD, AIDE, and DRCT (mAP) as well as DDA, DGS-Net, LTD, and PLM (mAP and Acc) on HiRes-50K (macro-averaged across the eight resolution intervals, matching RGE). Where a released checkpoint overlaps with published metrics, our evaluation agrees with the published scores up to small decimal differences. HiDA-Net has not released weights, so its unreported metrics are listed as N/A.

\section{Detailed Methodology}
\label{app:detailed-methodology}

This section details how F-lens decomposes backbone activations, selects the forensic reserve, and constructs the orthonormal writing bases for FRA, following Section~\ref{sec:method}.

\subsection{Component Probing and Reserve Identification}
\label{app:component-probing}

\noindent\textbf{Representation Collection.}
Let $\mathcal D_{\mathrm{disc}}=\{(x_i,y_i)\}_{i=1}^{N}$ denote the discovery fitting set, containing $N_y$ images in class $y\in\{0,1\}$. We record post-block activations from the frozen backbone at every candidate site $c=(\ell,t)$, where $\ell$ indexes the transformer block and $t$ indexes the token group (CLS, register, or patch). At each block, we keep all CLS and register tokens when present. For patch tokens, we partition the spatial grid into four quadrants and deterministically sample one token per quadrant (seeded by the image ID, block index, and sampling seed).

Let $\mathcal P_c(x_i)$ be the set of sampled token positions for image $x_i$ at site $c$, with count $q_{i,c}=|\mathcal P_c(x_i)|$. Each image carries a class-balanced weight $\omega_i=1/(2N_{y_i})$, which is split evenly across its sampled tokens as $\omega_i/q_{i,c}$. Summing over the sampled tokens at a site recovers the image weight $\omega_i$. Because the discovery pool is class-balanced ($N_0=N_1$) and $q_{i,c}$ is constant across images at a given site, every captured token row within site $c$ receives equal weight.

\noindent\textbf{Component Coordinates.}
Let $h^0_{c,p}(x_i)\in\mathbb R^{d_c}$ denote the unadapted activation of token $p\in\mathcal P_c(x_i)$ at site $c$. We apply $\ell_2$ normalization to remove magnitude variations:
\begin{equation}
\widehat h^0_{c,p}(x_i)
=\frac{h^0_{c,p}(x_i)}
{\max\!\left(\|h^0_{c,p}(x_i)\|_2,\epsilon_{\mathrm{norm}}\right)},
\label{eq:detail-normalization}
\end{equation}
with $\epsilon_{\mathrm{norm}}=10^{-12}$. Subtracting the site mean $\mu_c$ over the $n_c$ captured rows gives the centered matrix $H_c\in\mathbb R^{n_c\times d_c}$.

At each site $c$ with $\kappa_c\geq 1$, we set the number of ICA components using the rank of the centered data:
\begin{equation}
m_c
=\max\!\left\{
1,\,
\min\!\left(d_c,\ n_c-1,\ \kappa_c\right)
\right\},
\label{eq:detail-component-count}
\end{equation}
where $\kappa_c$ counts the eigenvalues of the centered Gram matrix exceeding $10\epsilon_{32}$ (float32 machine epsilon $\epsilon_{32}$), using whichever of $H_c^\top H_c$ or $H_cH_c^\top$ has smaller dimension.

Let $V_c\in\mathbb R^{d_c\times m_c}$ contain the retained eigenvectors in hidden-state space with eigenvalues $\lambda_c$ (when eigendecomposition uses $H_cH_c^\top$, its eigenvectors are first mapped back to hidden-state space). The whitening matrix is $Q_c=\operatorname{diag}(\lambda_c^{-1/2})V_c^\top$. We run parallel FastICA on $\sqrt{n_c}\,Q_cH_c^\top$ using the \texttt{logcosh} nonlinearity (up to 2,000 iterations, tolerance $10^{-4}$) to fit the unmixing matrix $W_c\in\mathbb R^{m_c\times m_c}$.

To place all components on a common scale before global ranking, let $\Gamma_c$ be the diagonal matrix of population standard deviations of the provisional responses $W_cQ_c(\widehat h^0_{c,p}-\mu_c)$ across the captured rows. Dividing by $\Gamma_c$ gives unit-variance component coordinates:
\begin{equation}
z_{c,p}(x)
=\Gamma_c^{-1}W_cQ_c
\left(\widehat h^0_{c,p}(x)-\mu_c\right)
=R_c\left(\widehat h^0_{c,p}(x)-\mu_c\right),
\label{eq:detail-component-coordinates}
\end{equation}
where $R_c=\Gamma_c^{-1}W_cQ_c\in\mathbb R^{m_c\times d_c}$ is the final reading map, carrying the same scaling into its synthesis directions.

\noindent\textbf{Sensitivity and Selection.}
The class-conditional mean response of component $j$ at site $c$ in Eq.~(\ref{eq:origin-sensitivity}) is
\begin{equation}
\overline z^{(y)}_{c,j}
=\frac{
\displaystyle\sum_{i:y_i=y}
\frac{\omega_i}{q_{i,c}}
\sum_{p\in\mathcal P_c(x_i)}z_{c,p,j}(x_i)
}{\displaystyle\sum_{i:y_i=y}\omega_i},
\label{eq:detail-class-response}
\end{equation}
which equals the sample mean over class-$y$ token rows at site $c$. The origin sensitivity is $s_{c,j}=|\overline z^{(1)}_{c,j}-\overline z^{(0)}_{c,j}|$.

Pooling all fitted sites yields $M=\sum_c m_c$ candidate components. For selection fraction $\rho$ ($\rho=0.01$ by default), we rank all $M$ candidates by decreasing $s_{c,j}$ (with deterministic tie-breaking by site identifier and component index) and retain the top $K=\max(1,\lceil\rho M\rceil)$ components. For each site $c$, let $\mathcal I_c$ contain its retained component indices. The global ranking thus determines both the number of components retained at each site and the set of active sites $\mathcal C_\star=\{c:\mathcal I_c\neq\varnothing\}$ in Section~\ref{sec:reserve-localization}.

\subsection{Writing Subspaces and FRA Optimization}
\label{app:writing-optimization}

\noindent\textbf{Synthesis Directions and Writing Bases.}
Following the reading and synthesis construction of~\citet{liu2026icalens}, we map component coordinates back to hidden-state space through the synthesis matrix $D_c\in\mathbb R^{d_c\times m_c}$:
\begin{equation}
D_c=R_c^\dagger,
\label{eq:detail-synthesis}
\end{equation}
where $\dagger$ is the Moore--Penrose pseudoinverse (float64 SVD, retaining singular values above $10^{-12}\sigma_{\max}(R_c)$).

At each active site $c\in\mathcal C_\star$, the columns $B_c=D_c[:,\mathcal I_c]$ span the mapped reserve subspace $D_cF_c=\operatorname{span}(B_c)$. We compute the float64 SVD of $B_c$ and retain the left singular vectors whose singular values exceed $\max(d_c,|\mathcal I_c|)\epsilon_{64}\sigma_{\max}(B_c)$. These vectors form the orthonormal writing basis $U_c\in\mathbb R^{d_c\times r_c}$ in Eq.~(\ref{eq:reserve-writing-space}), where $r_c$ is the numerical rank of $B_c$.

\noindent\textbf{Reference Detector and FRA Optimization.}
The reference classifier and feature standardization follow Section~\ref{sec:reserve-elicitation}; any feature dimension with zero empirical standard deviation is assigned unit scale. During FRA training, the class-balanced sample weights $\omega_i$ in Eq.~(\ref{eq:detail-detection-loss}) are implemented by drawing mini-batches uniformly with replacement from the balanced fitting pool and minimizing the batch-mean binary cross-entropy. Gradients backpropagate through the frozen scoring head and backbone to update only the FRA coefficient maps $\mathcal A=\{A_c:c\in\mathcal C_\star\}$.

\section{Additional Analyses}
\label{app:additional-analyses}

\begin{figure*}[p]
\centering
\includegraphics[width=0.98\linewidth]{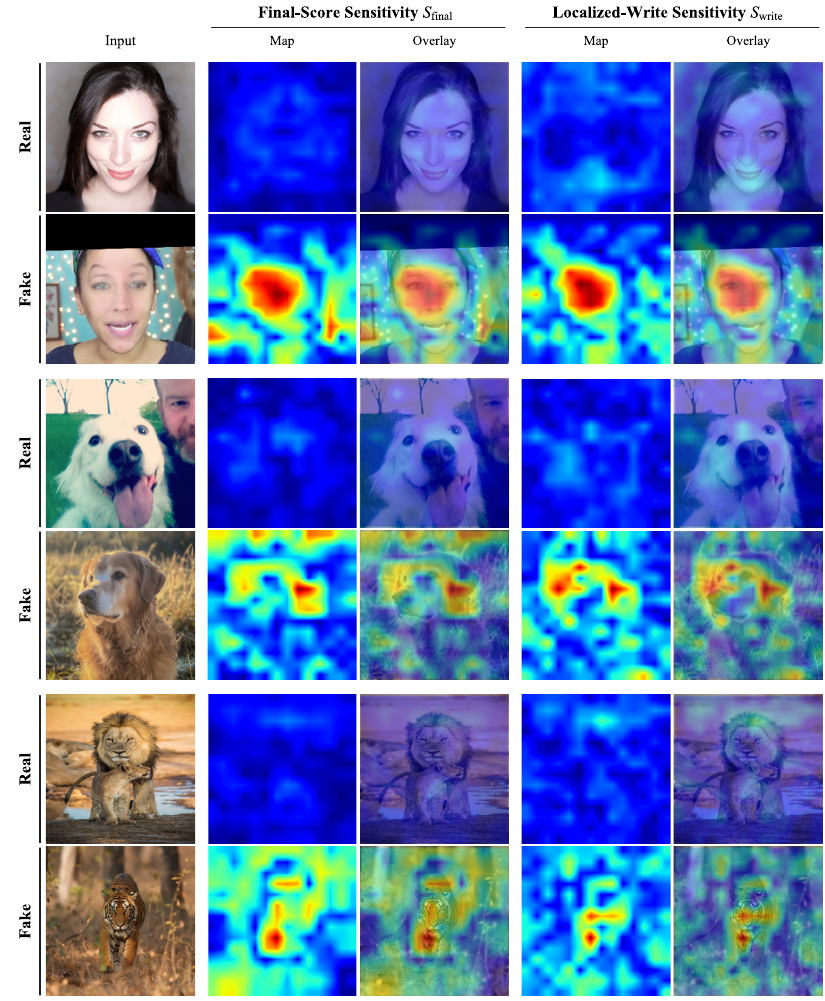}
\caption{\textbf{Final-Score and Localized-Write Sensitivity.} $S_{\mathrm{final}}$ and $S_{\mathrm{write}}$ for Figure~\ref{fig:fra-spatial}, normalized by map type within each pair.}
\label{fig:fra-spatial-write}
\end{figure*}
\subsection{Functional Analyses}
\label{app:additional-functional}

\noindent\textbf{AIGIBench Analysis Panel.}
Unless noted otherwise, all post-hoc interventions, write-gate sweeps, component-response plots, and perturbation tests on the 5K RGE detector are evaluated on a fixed, class-balanced panel of 2,000 AIGIBench images (40 real and 40 fake images sampled with a fixed seed from each of the 25 AIGIBench subsets).

\noindent\textbf{Analysis Quantities.}
To measure how the learned writes affect predictions without retraining, we scale each site's residual update in Eq.~(\ref{eq:fra-update}) by a gate $g_c\in[0,1]$, giving $h'_{c,p}=h_{c,p}+g_cU_cA_ch_{c,p}$. Let $f_{\mathcal A}(x;\mathbf g)$ denote the detector score under gate vector $\mathbf g$. Setting $\mathbf g=\mathbf 0$ recovers the frozen reference detector, $\mathbf g=\mathbf 1$ gives the trained RGE detector, and a global gate sets $g_c=g$ at all sites.
\emph{Gate IG.} We evaluate Eq.~(\ref{eq:main-gate-ig}) (with $f=f_{\mathcal A}$) via the midpoint rule along the straight path $\mathbf g=t\mathbf 1$ for $t\in[0,1]$. Multiplying by $(2y-1)$ aligns the sign with the true label so that positive values mark writes that push the score toward the correct class.
\emph{Probe AP.} At each active site, we average the site's reserve-component responses across tokens to form a site representation for each discovery image, then compute five-fold out-of-fold average precision using the same $\ell_2$-regularized logistic probe as the reference classifier (folds split by hashed image ID). Probe AP thus quantifies how strongly each site's selected components separate real and fake images on the discovery set. We then test whether this discovery-set site ranking transfers to held-out data via the site knockout experiment below.
\emph{Site Knockout.} Let $\mathcal H$ be the six sites with the highest discovery-set probe AP. On the held-out AIGIBench analysis panel, we compare disabling $\mathcal H$ ($g_c=0$ for $c\in\mathcal H$) with disabling three random six-site groups of approximately matched total writing rank.

\noindent\textbf{Responses in a Fixed ICA Coordinate.}
\label{app:component-separation}
Figure~\ref{fig:component-separation}~(b) evaluates a single discovery-selected reserve component across the 2,000-image analysis panel with all FRA writes disabled ($g=0$) versus enabled ($g=1$). In both conditions, we project the block output onto the same discovery-fitted ICA reading vector with fixed centering and sign, so the shift between the two distributions reflects the effect of the FRA writes (both upstream and at that block). The Gaussian kernel density curves use a single shared bandwidth across classes and gate states (the root-mean-square of the four within-group sample standard deviations multiplied by $1000^{-1/5}$), with shared axis limits and no sample clipping.

\noindent\textbf{Image Regions and Localized Writes.}
For a patch grid location $u$, let $x_u$ denote image $x$ after blurring the $3\times 3$ patch window centered at $u$ (three passes of a 9-pixel box filter). Final-score sensitivity is $S_{\mathrm{final}}(u)=f_{\mathcal A}(x;\mathbf 1)-f_{\mathcal A}(x_u;\mathbf 1)$. To isolate the spatial contribution of the top-six reserve sites $\mathcal H$, let $G(x)=f_{\mathcal A}(x;\mathbf 1)-f_{\mathcal A}(x;\mathbf 1_{\setminus\mathcal H})$ be their net score contribution, where $\mathbf 1_{\setminus\mathcal H}$ disables only $\mathcal H$, and define localized-write sensitivity as $S_{\mathrm{write}}(u)=G(x)-G(x_u)$. Figure~\ref{fig:fra-spatial-write} compares $|S_{\mathrm{final}}|$ from Figure~\ref{fig:fra-spatial} with $|S_{\mathrm{write}}|$. In all three fake images, both sensitivity magnitudes are highest over the same synthesized subject regions (the portrait face, the dog's head, and the tiger's face and foreleg), while both remain low in the real images. Across all three pairs, the top-six reserve writes contribute to the detector's sensitivity in the same subject regions that drive its final score. To render the maps, absolute values on the $14\times 14$ patch grid are normalized by the shared maximum within each real/fake pair (separately for $S_{\mathrm{final}}$ and $S_{\mathrm{write}}$), upsampled via bilinear interpolation, mapped through a power-law color scale ($\gamma=0.7$) from blue (low) to red (high), and blended equally with the input image.

\subsection{Adaptation Analyses}
\label{app:additional-adaptation}

\noindent\textbf{Writing Capacity.}
Table~\ref{tab:reserve-capacity} lists total writing ranks and trainable parameter counts for six selection fractions across the 0.5K, 5K, and 20K discovery pools. In DINOv3-H+ ($d_c=1{,}280$), each retained basis direction introduces 1,280 trainable coefficients. At the default 1\% selection budget, the total rank ranges from 918 to 1,196 across the three discovery sizes, corresponding to 1.18M--1.53M trainable parameters (0.14\%--0.18\% of the 840.6M-parameter backbone). At this budget, selected components span most layer--token sites with small per-site ranks, so the reserve is sparse within sites rather than across sites. Instead of concentrating capacity in a few layers, F-lens assigns each site a low-dimensional subspace aligned with its local origin-sensitive directions. Disregarding this cross-token allocation by shifting the CLS budget to patch and register sites lowers AIGIBench mAP by 4.57 points (Table~\ref{tab:structure-allocation}~(b)).

\begin{table*}[t]
\centering
\caption{\textbf{Reserve Selection and Trainable Capacity.} Six retained component fractions across three discovery sizes. The shaded row is the default top-1\% configuration. The backbone denominator is 840.6M parameters.}
\label{tab:reserve-capacity}
\begingroup
\fontsize{8}{9.5}\selectfont
\setlength{\tabcolsep}{2pt}
\renewcommand{\arraystretch}{1.16}
\begin{tabularx}{\linewidth}{@{}r*{9}{>{\raggedleft\arraybackslash}X}@{}}
\toprule
& \multicolumn{3}{c}{\textbf{0.5K Discovery}} & \multicolumn{3}{c}{\textbf{5K Discovery}} & \multicolumn{3}{c}{\textbf{20K Discovery}} \\
\cmidrule(lr){2-4}\cmidrule(lr){5-7}\cmidrule(lr){8-10}
\textbf{Fraction} & Rank & Parameters & Backbone & Rank & Parameters & Backbone & Rank & Parameters & Backbone \\
\midrule
0.1\% & 92 & 117,760 & 0.0140\% & 118 & 151,040 & 0.0180\% & 120 & 153,600 & 0.0183\% \\
0.25\% & 230 & 294,400 & 0.0350\% & 294 & 376,320 & 0.0448\% & 299 & 382,720 & 0.0455\% \\
0.5\% & 459 & 587,520 & 0.0699\% & 587 & 751,360 & 0.0894\% & 598 & 765,440 & 0.0911\% \\
\rowcolor{rowTint}1\% & 918 & 1,175,040 & 0.1398\% & 1,174 & 1,502,720 & 0.1788\% & 1,196 & 1,530,880 & 0.1821\% \\
2\% & 1,835 & 2,348,800 & 0.2794\% & 2,347 & 3,004,160 & 0.3574\% & 2,391 & 3,060,480 & 0.3641\% \\
5\% & 4,588 & 5,872,640 & 0.6986\% & 5,867 & 7,509,760 & 0.8934\% & 5,978 & 7,651,840 & 0.9103\% \\
\bottomrule
\end{tabularx}
\endgroup
\end{table*}

\section{Additional Results}
\label{app:additional-results}

\begin{figure*}[t]
\centering
\includegraphics[width=0.98\linewidth]{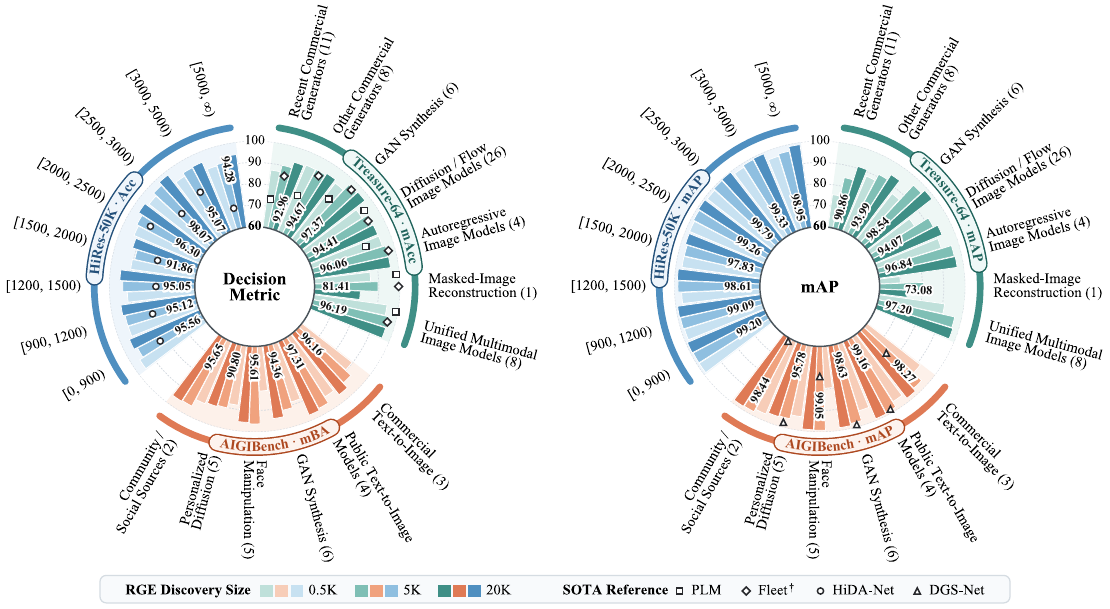}

\caption{\textbf{Detection by Source Category and Resolution Interval (\%).} Decision metric (left; mAcc, mBA, and Acc) and mAP (right) of RGE at 0.5K, 5K, and 20K discovery images (light to dark; labels give 20K), with published references as markers. Parentheses give source counts; HiRes-50K intervals give the image long edge in pixels. Bars start at 60\%. Fleet\textsuperscript{$\dagger$} uses 10-shot adaptation on Treasure-64.}
\label{fig:source-radial}
\end{figure*}
\subsection{Benchmark Source Breakdown}
\label{app:source-breakdown}
\noindent\textbf{Category Aggregation and Main Trends.}
Figure~\ref{fig:source-radial} shows how RGE performance varies across AIGIBench source categories and HiRes-50K resolution intervals at 0.5K, 5K, and 20K discovery budgets, with decision metrics (mBA and Acc) evaluated at each detector's development-set threshold. Category scores preserve native subset weights, so weighting category means by their subset counts (in parentheses) recovers the overall benchmark score.

Figure~\ref{fig:source-radial} highlights two trends on AIGIBench and HiRes-50K. First, on AIGIBench, the 20K detector reaches 94.36\%--97.31\% mBA across five of the six categories: public text-to-image models (97.31\% mBA), commercial text-to-image models (96.16\%), community and social sources (95.65\%), face manipulation (95.61\%), and GAN synthesis (94.36\%). Across these five categories, RGE's category mAP reaches 98.27\%--99.16\%, outperforming DGS-Net by wide margins on commercial text-to-image models, face manipulation, and community and social sources. Personalized diffusion is the hardest category on AIGIBench (90.80\% mBA, 95.78\% mAP), which is consistent with reference-conditioned methods preserving visual characteristics of the real reference images. Second, on HiRes-50K, 20K RGE maintains high detection performance across all eight native-resolution bands (91.86\%--98.07\% Acc and 97.83\%--99.79\% mAP), from below 900 pixels (95.56\% Acc, 99.20\% mAP) to at least 5,000 pixels (94.28\% Acc, 98.95\% mAP), complementing the resizing robustness results in Figure~\ref{fig:fra-robustness-gain}.

\noindent\textbf{AIGIBench Categories.}
We group the 25 native subsets of AIGIBench~\citep{li2025aigibench} into six categories.
\emph{Commercial text-to-image}: DALLE-3, Imagen3, Midjourney.
\emph{Public text-to-image models}: FLUX1-dev, GLIDE~\generatorcite{GLIDE}, SD3, SDXL~\generatorcite{SDXL}.
\emph{GAN synthesis}: ProGAN~\generatorcite{ProGAN}, R3GAN, StyleGAN3~\generatorcite{StyleGAN3}, StyleGAN-XL, StyleSwin~\generatorcite{StyleSwin}, WFIR.
\emph{Face manipulation}: BlendFace, E4S, FaceSwap, InSwap, SimSwap.
\emph{Personalized diffusion}: BLIP~\generatorcite{BLIP}, Infinite-ID, InstantID, IP-Adapter, PhotoMaker~\generatorcite{PhotoMaker}.
\emph{Community / social sources}: the CommunityAI and SocialRF subsets.

\begin{table}[!t]
\centering
\caption{\textbf{Detection on Treasure-64 (\%).} RGE uses 5K discovery images; superscripts give its change from the baseline. Best and second-best results are bold and underlined.}
\label{tab:treasure-results}
\begingroup
\small
\setlength{\tabcolsep}{2pt}
\renewcommand{\arraystretch}{1.21}
\newcommand{\dup}[1]{\rlap{\textsuperscript{\,\fontsize{5.5}{6}\selectfont\textcolor{deltaGreen}{+#1}}}}
\begin{tabularx}{\linewidth}{@{}>{\raggedright\arraybackslash}X>{\raggedright\arraybackslash\hspace*{14.15pt}}p{48.5pt}>{\raggedright\arraybackslash\hspace*{13.12pt}}p{48.5pt}@{}}
\toprule
\textbf{Method} & \multicolumn{1}{c}{mAP $\uparrow$} & \multicolumn{1}{c}{mAcc $\uparrow$} \\
\midrule
FreqNet~\citep{tan2024freqnet} & 77.71 & 73.42 \\
UnivFD~\citep{ojha2023univfd} & 84.33 & 74.17 \\
AIDE~\citep{yan2025aide} & 83.72 & 84.75 \\
SAFE~\citep{li2025safe} & 86.26 & 86.66 \\
PLM~\citep{zhou2026plm} & 90.44 & 88.18 \\
FSD (10-shot)~\citep{wu2025fsd} & 80.54 & 75.87 \\
Fleet (10-shot)~\citep{wang2026fleet} & \textbf{95.66} & \underline{92.63} \\
\midrule
Baseline & 81.66 & 85.92 \\
\rowcolor{oursLav}\textbf{RGE (Ours)} & \underline{91.65}\dup{9.99} & \textbf{93.85}\dup{7.93} \\
\bottomrule
\end{tabularx}
\endgroup
\end{table}
\begin{table*}[t]
\centering
\caption{\textbf{Generator Families in Treasure-64.} Generator versions follow Treasure-64~\citep{wang2026fleet}, which also lists the image sources.}
\label{tab:treasure-families}
\fontsize{8.5}{10}\selectfont
\setlength{\tabcolsep}{4pt}
\renewcommand{\arraystretch}{1.13}
\begin{tabularx}{\linewidth}{@{}l>{\raggedright\arraybackslash}X@{}}
\toprule
\textbf{Category (count)} & \textbf{Generators} \\
\midrule
\textbf{Recent Commercial Generators (11)} & \textit{\mbox{GPT4O\_Image\_T2I~\generatorcite{treasure27}}, \mbox{Imagen4~\generatorcite{treasure51}}, \mbox{Nano Banana~\generatorcite{treasure52}}, \mbox{Nano-Banana-Pro~\generatorcite{treasure53}}, \mbox{doubao-seedream-3.0-t2i~\generatorcite{treasure55}}, \mbox{doubao-seedream-4.0~\generatorcite{treasure54}}, \mbox{wan2.2-t2i-flash~\generatorcite{treasure58}}, \mbox{wan2.5-t2i-preview~\generatorcite{treasure58}}, \mbox{sora-image~\generatorcite{treasure27}}, \mbox{gpt-image-1.5~\generatorcite{treasure62}}, \mbox{Midjourney V7~\generatorcite{treasure64}}} \\\addlinespace[7.5pt]
\textbf{Other Commercial Generators (8)} & \textit{\mbox{DALLE-2~\generatorcite{treasure13}}, \mbox{DALLE-3~\generatorcite{treasure24}}, \mbox{Imagen~\generatorcite{treasure14}}, \mbox{Midjourney V4~\generatorcite{treasure64}}, \mbox{Midjourney V5~\generatorcite{treasure64}}, \mbox{Midjourney V6~\generatorcite{treasure64}}, \mbox{Midjourney V6.1~\generatorcite{treasure64}}, \mbox{ideogram~\generatorcite{treasure22}}} \\\addlinespace[7.5pt]
\textbf{GAN Synthesis (6)} & \textit{\mbox{BigGAN~\generatorcite{BigGAN}}, \mbox{DF-GAN~\generatorcite{DF-GAN}}, \mbox{GigaGAN~\generatorcite{GigaGAN}}, \mbox{ProGAN~\generatorcite{ProGAN}}, \mbox{StarGAN~\generatorcite{StarGAN}}, \mbox{StyleGAN3~\generatorcite{StyleGAN3}}} \\\addlinespace[7.5pt]
\textbf{Diffusion / Flow Image Models (26)} & \textit{\mbox{ADM~\generatorcite{treasure2}}, \mbox{BRIA\_v3\_2~\generatorcite{treasure42}}, \mbox{CogView3-plus~\generatorcite{treasure35}}, \mbox{DeepFloyd\_IF~\generatorcite{treasure21}}, \mbox{FLUX.1-dev~\generatorcite{treasure25}}, \mbox{GLIDE~\generatorcite{GLIDE}}, \mbox{HiDream-I1-Dev~\generatorcite{treasure40}}, \mbox{HunyuanDiT~\generatorcite{treasure30}}, \mbox{LongCat-Image~\generatorcite{treasure48}}, \mbox{Lumina~\generatorcite{treasure39}}, \mbox{Playground\_v2~\generatorcite{treasure28}}, \mbox{Playground\_v2.5~\generatorcite{treasure29}}, \mbox{SD3-Medium~\generatorcite{treasure32}}, \mbox{SDXL~\generatorcite{SDXL}}, \mbox{SDv1.4~\generatorcite{treasure6}}, \mbox{SDv1.5~\generatorcite{treasure6}}, \mbox{SDv2.1~\generatorcite{treasure20}}, \mbox{Sana\_v1.5~\generatorcite{treasure38}}, \mbox{VQDM~\generatorcite{treasure5}}, \mbox{Wukong~\generatorcite{treasure4}}, \mbox{Z-Image-Turbo~\generatorcite{treasure47}}, \mbox{pixart-$\alpha$~\generatorcite{PixArt}}, \mbox{Kolors~\generatorcite{treasure49}}, \mbox{Qwen-Image~\generatorcite{treasure50}}, \mbox{FLUX.2~\generatorcite{treasure57}}, \mbox{CogView4~\generatorcite{treasure60}}} \\\addlinespace[7.5pt]
\textbf{Autoregressive Image Models (4)} & \textit{\mbox{CogView2~\generatorcite{CogView2}}, \mbox{Infinity~\generatorcite{treasure36}}, \mbox{LlamaGen~\generatorcite{treasure31}}, \mbox{NextStep~\generatorcite{treasure46}}} \\\addlinespace[7.5pt]
\textbf{Masked-image Reconstruction (1)} & \textit{\mbox{MAE~\generatorcite{MAE}}} \\\addlinespace[7.5pt]
\textbf{Unified Multimodal Image Models (8)} & \textit{\mbox{BAGEL-7B~\generatorcite{treasure41}}, \mbox{Janus-Pro-7B~\generatorcite{treasure37}}, \mbox{OmniGen\_v1~\generatorcite{treasure34}}, \mbox{OmniGen\_v2~\generatorcite{treasure43}}, \mbox{Show\_o~\generatorcite{treasure33}}, \mbox{Show\_o2~\generatorcite{treasure44}}, \mbox{ovis-U1~\generatorcite{treasure45}}, \mbox{HunyuanImage-3.0~\generatorcite{treasure56}}} \\
\bottomrule
\end{tabularx}
\end{table*}
\noindent\textbf{HiRes-50K, Chameleon, and Reference Markers.}
HiRes-50K defines eight native long-edge resolution intervals in pixels: $[0,900)$, $[900,1200)$, $[1200,1500)$, $[1500,2000)$, $[2000,2500)$, $[2500,3000)$, $[3000,5000)$, and $[5000,\infty)$, whose unweighted mean gives the benchmark macro score~\citep{mu2026hida}. Because the Chameleon benchmark reports only overall and per-class accuracy without a generator breakdown~\citep{yan2025aide}, its results are reported only in Table~\ref{tab:main-results}. Reference markers in Figure~\ref{fig:source-radial} show published source-level scores where available: category AP on AIGIBench~\citep{li2025aigibench} comes from per-source AP reported for DGS-Net~\citep{yan2026dgsnet}, and HiDA-Net interval Acc comes from~\citet{mu2026hida}. DDA does not report source-level results~\citep{chen2025dda}.

\subsection{Evaluation on Treasure-64}
\label{app:treasure-evaluation}

\noindent\textbf{Benchmark and Evaluation Setup.}
We further evaluate RGE on Treasure-64~\citep{wang2026fleet}, which tests generalization across 64 image generators using mAP and mean accuracy (mAcc). FreqNet~\citep{tan2024freqnet}, UnivFD, AIDE, SAFE, and PLM are evaluated without target-benchmark adaptation~\citep{wang2026fleet,zhou2026plm}. In contrast, FSD~\citep{wu2025fsd} and Fleet perform 10-shot adaptation using labeled images from Treasure-64 (Fleet first pretrains LoRA adapters and frequency-guided subspace routing on DINOv3 with 144K images, then adapts its routing with 10-shot Treasure-64 samples)~\citep{wang2026fleet}. RGE in Table~\ref{tab:treasure-results} uses 5K discovery images without target-benchmark adaptation.

\noindent\textbf{Overall and Category-Level Results.}
As shown in Table~\ref{tab:treasure-results}, 5K RGE improves over its frozen baseline by 9.99 mAP and 7.93 mAcc points. Among methods evaluated without target-benchmark adaptation, RGE achieves the highest mAP (91.65\% vs.\ 90.44\% for PLM) and mAcc (93.85\% vs.\ 88.18\% for PLM). Compared with 10-shot Fleet, RGE attains higher mAcc (93.85\% vs.\ 92.63\%) while trailing in mAP (91.65\% vs.\ 95.66\%).

Across the seven generator families in Figure~\ref{fig:source-radial}, 20K RGE maintains high accuracy across six distinct generator families: 92.96\% mAcc on recent commercial generators (11 models, including GPT-4o, Imagen4, and Midjourney V7), 94.67\% on other commercial generators (8), 97.37\% on GANs (6), 94.41\% on diffusion and flow models (26), 96.06\% on autoregressive models (4), and 96.19\% on unified multimodal models (8). Masked-image reconstruction (MAE, 1 model) is the main exception, reaching 81.41\% mAcc and 73.08\% mAP at 20K and scoring lower at 20K than at 5K. MAE~\generatorcite{MAE} reconstructs masked patches from the visible pixels of a real image. Preserving these pixels may leave fewer image-wide clues to synthesis.

\noindent\textbf{Source Categories and Aggregation.}
Table~\ref{tab:treasure-families} groups the 64 generators into seven families.
Category mAP averages the constituent generator APs. Category mAcc averages $(\mathrm{TPR}_g+\mathrm{TNR}_{\mathrm{real}})/2$ over its generators, where $\mathrm{TNR}_{\mathrm{real}}$ is the pooled real-image true-negative rate across COCO2014 and cc12m-2mp-realistic (excluded from the 64-generator AP average). PLM and Fleet\textsuperscript{$\dagger$} category mAcc markers in Figure~\ref{fig:source-radial} use per-generator accuracies from the Fleet paper~\citep{wang2026fleet}, which does not report per-generator AP.

\section{Limitations and Future Work}
\label{app:limitations}
Our evaluation covers still images from the generators represented in our benchmarks. As new proprietary image generators are released, evaluating RGE and its frozen reference detector on images from these systems, with both detectors held fixed, would show how the gain from reserve-guided adaptation changes over time. Video forensics offers a complementary direction. Applying F-lens to video representations would allow us to examine whether motion and other temporal cues carry evidence of manipulation beyond what is visible in individual frames.

\subsection*{Ethics Statement}
This study uses only publicly released datasets and open-weight pretrained models and collects no new personal data. Source datasets containing human faces are used under their original release terms and are not redistributed. Like other learning-based detectors, RGE can reflect biases present in its discovery sources and pretrained backbones, and detection accuracy can vary across generators, content domains, and image quality. Localizing internal forensic representations also carries dual-use risks, since understanding which subspaces drive detection could assist adversarial evasion. RGE is designed to support human review. Before deployment, its detection performance should be validated on images representative of the intended domain, and its use in practice should remain subject to human oversight.

\twocolumn[{\captionof{table}{\textbf{Notation for Reserve Localization, Adaptation, and Gate Analysis.}}
\label{tab:reserve-notation}
\begingroup
\small
\renewcommand{\arraystretch}{1.18}
\begin{tabularx}{\textwidth}{@{}l>{\raggedright\arraybackslash}X@{}}
\toprule
\textbf{Symbol} & \textbf{Meaning} \\
\midrule
$\mathcal D$, $\mathcal D_{\mathrm{disc}}$ & Discovery fitting set, denoted by $\mathcal D$ in Algorithm~\ref{alg:rge} and $\mathcal D_{\mathrm{disc}}$ in Appendix~\ref{app:component-probing}. \\
$\mathcal D_{\mathrm{dev}}$ & Held-out development set used to select the reference threshold $\tau$. \\
$c=(\ell,t)$ & Candidate site indexed by transformer block $\ell$ and token group $t$. \\
$p$, $j$, $d_c$ & Token position, component index, and hidden-state dimension at site $c$. \\
$h^0_{c,p}$, $\widehat h^0_{c,p}$, $\mu_c$ & Unadapted token state, its $\ell_2$-normalized form, and the site mean of normalized states. \\
$z_{c,p}$, $s_{c,j}$ & Component coordinates and origin sensitivity: $s_{c,j}$ is the absolute difference in class means of coordinate $j$. \\
$m_c$, $M$ & Number of candidate components at site $c$ and total across fitted sites, $M=\sum_c m_c$. \\
$\rho$, $K$ & Selection fraction and global number of retained components, $K=\max(1,\lceil\rho M\rceil)$. \\
$\mathcal I_c$, $\mathcal C_\star$ & Indices retained at site $c$ and set of active sites, $\mathcal C_\star=\{c:\mathcal I_c\ne\varnothing\}$. \\
\addlinespace[3pt]
$R_c$, $D_c$ & Reading map into component coordinates and synthesis map $D_c=R_c^\dagger$ back to hidden space. \\
$F_c$, $U_c$ & Selected coordinate subspace at site $c$ and orthonormal writing basis spanning its image $D_c F_c$ in hidden space. \\
$r_c$ & Numerical rank of $D_c[:,\mathcal I_c]$, equal to the number of columns in $U_c$. \\
$A_c$, $\mathcal A$ & Trainable coefficient matrix at site $c$ and collection of these matrices over active sites, $\mathcal A=\{A_c:c\in\mathcal C_\star\}$. \\
$h_{c,p}$, $h'_{c,p}$ & Current and updated token states in FRA: $h'_{c,p}=h_{c,p}+U_c A_c h_{c,p}$. \\
$\Phi_{\theta,\mathcal A}(x)$, $\psi$, $f_{\mathcal A}(x)$ & Standardized pooled feature with frozen backbone parameters $\theta$; the frozen head $\psi(v)=w^\top v+b-\tau$ gives the logit $f_{\mathcal A}(x)=\psi(\Phi_{\theta,\mathcal A}(x))$. \\
$\mathcal L(\mathcal A)$, $\omega_i$ & Weighted binary cross-entropy for fitting $\mathcal A$, with $\omega_i=1/(2N_{y_i})$ for an image in class $y_i$ of size $N_{y_i}$. \\
$g_c$, $\mathbf g$ & Site gate and gate vector used in functional analyses. $g_c=0$ disables the write at site $c$. \\
$\mathrm{IG}_c(x)$ & Site $c$'s contribution to the logit change from reference to adapted detector for image $x$, integrated along the gate path $\mathbf g=t\mathbf 1$, $0\le t\le 1$. \\
\bottomrule
\end{tabularx}
\endgroup
\par\bigskip}]
\end{document}